%% file: main.tex
\def\isarxiv{1}

\ifdefined\isarxiv
\documentclass[11pt]{article}
\else

\relax
\documentclass[letterpaper]{article} 
\usepackage{aaai22}  
\usepackage{times}  
\usepackage{helvet}  
\usepackage{courier}  
\usepackage[hyphens]{url}  
\usepackage{graphicx} 
\usepackage{natbib}  
\usepackage{caption} 
\DeclareCaptionStyle{ruled}{labelfont=normalfont,labelsep=colon,strut=off} 
\usepackage{algorithm}

\usepackage{newfloat}
\usepackage{listings}

\floatstyle{ruled}
\newfloat{listing}{tb}{lst}{}
\floatname{listing}{Listing}
\fi

\usepackage{microtype}
\usepackage{amsmath}
\usepackage{amsthm}
\allowdisplaybreaks
\usepackage{amssymb}
\usepackage{subfig}
\usepackage{color}
\usepackage{epstopdf}
\usepackage{algpseudocode}
\usepackage{scrextend}
\usepackage[T1]{fontenc}
\usepackage{bbm}
\usepackage{comment}
\usepackage{booktabs}

\usepackage{multicol}
\usepackage{dsfont}
\usepackage{mathtools}
\usepackage{enumitem}

\usepackage{tikz}

\usetikzlibrary{arrows}
\ifdefined\isarxiv
\usepackage[margin=1in]{geometry}
\usepackage{hyperref}
\hypersetup{colorlinks=true,citecolor=red,linkcolor=red}
\else

\fi
\graphicspath{{./figs/}}

\definecolor{b2}{RGB}{51,153,255}
\definecolor{mygreen}{RGB}{80,180,0}
\definecolor{yl}{RGB}{255,80,0}

\newcommand{\Zhao}[1]{\textcolor{b2}{[Zhao: #1]}}
\newcommand{\Yunze}[1]{\textcolor{red}{[Yunze: #1]}}
\newcommand{\Danyang}[1]{\textcolor{mygreen}{[Danyang: #1]}}
\newcommand{\Shunhua}[1]{{\color{yl}[Shunhua: #1]}}
\newcommand{\blue}[1]{{\color{blue} #1}}

\theoremstyle{plain}
\newtheorem{theorem}{Theorem}[section]
\newtheorem{lemma}[theorem]{Lemma}

\newtheorem{fact}[theorem]{Fact}

\newtheorem{assumption}[theorem]{Assumption}

\newtheorem{definition}[theorem]{Definition}

\theoremstyle{remark}

\DeclareMathOperator*{\E}{{\mathbb{E}}}

\graphicspath{{./figs/}}

\newcommand{\wt}{\widetilde}
\newcommand{\ov}{\overline}

\renewcommand{\epsilon}{\varepsilon}
\newcommand{\N}{\mathcal{N}}
\newcommand{\R}{\mathbb{R}}

\renewcommand{\k}{\mathsf{K}}

\renewcommand{\d}{\mathrm{d}}
\newcommand{\poly}{\mathrm{poly}}

\newcommand{\vect}{\mathrm{vec}}
\newcommand{\Tmat}{{\cal T}_{\mathrm{mat}}}
\newcommand{\nn}{\mathrm{nn}}
\newcommand{\gnn}{\mathrm{gnn}}
\newcommand{\ntk}{\mathrm{ntk}}
\newcommand{\gntk}{\mathrm{gntk}}

\newcommand{\tr}{\mathrm{tr}}
\newcommand{\neighbor}{\mathcal{N}}
\newcommand{\block}{\textsc{Block}}
\newcommand{\aggregate}{\textsc{Aggregate}}
\newcommand{\combine}{\textsc{Combine}}
\newcommand{\readout}{\textsc{ReadOut}}

\makeatletter
\newcommand*{\RN}[1]{\expandafter\@slowromancap\romannumeral #1@}
\makeatother

\title{Fast Graph Neural Tangent Kernel via Kronecker Sketching\footnote{A preliminary version of this paper appeared in 36th AAAI Conference on Artificial Intelligence (AAAI 2022).}}

\ifdefined\isarxiv
\author {
    Shunhua Jiang\thanks{Columbia University. \texttt{sj3005@columbia.edu}.}
    \and
    Yunze Man\thanks{Carnegie Mellon University. \texttt{yman@cs.cmu.edu}.}
    \and 
    Zhao Song\thanks{Adobe Research. \texttt{zsong@adobe.com}.}
    \and 
    Zheng Yu\thanks{Princeton University. \texttt{zhengy@princeton.edu}.}
    \and
    Danyang Zhuo\thanks{Duke University. \texttt{danyang@cs.duke.edu}.}
}
   
\else

\author {
    Shunhua Jiang\thanks{Columbia University. \texttt{sj3005@columbia.edu}.}
    Yunze Man\thanks{Carnegie Mellon University. \texttt{yman@cs.cmu.edu}.}
    Zhao Song\thanks{Adobe Research. \texttt{zsong@adobe.com}.}
    Zheng Yu\thanks{Princeton University. \texttt{zhengy@princeton.edu}.}
    Danyang Zhuo\thanks{Duke University. \texttt{danyang@cs.duke.edu}.}
}
\fi

\usepackage{bibentry}

\ifdefined\isarxiv

\else
\fi

\begin{document}

\ifdefined\isarxiv
 \begin{titlepage}
     \maketitle
     \begin{abstract}
         \input{abstract}
     \end{abstract}
     \thispagestyle{empty}
 \end{titlepage}
 \newpage
\else
\maketitle
     \begin{abstract}
         \input{abstract}
     \end{abstract}
\fi


\input{intro}

\input{related}

\input{problem} 

\input{kronecker}
\input{conclusion}

\newpage

\addcontentsline{toc}{section}{References}
\ifdefined\isarxiv
\bibliographystyle{alpha}
\else

\fi
\bibliography{ref}

\newpage

\appendix

 \onecolumn
\input{app_preli}
\input{gntk_formula}
\input{gntk_generalization_bound}
\input{gntk_time}
\input{app_sketching}
\input{exp}

\end{document}

%% file: abstract.tex
Many deep learning tasks have to deal with graphs (e.g., protein structures, social networks, source code abstract syntax trees).
Due to the importance of these tasks, people turned to Graph Neural Networks (GNNs) as the de facto method for learning on graphs. GNNs have become widely applied due to their convincing performance. Unfortunately, one major barrier to using GNNs is that GNNs require substantial time and resources to train. 
Recently, a new method for learning on graph data is \textit{Graph Neural Tangent Kernel (GNTK)}~\cite{dhs+19}.
GNTK is an application of Neural Tangent Kernel (NTK)~\cite{jgh18} (a kernel method) on graph data, and solving NTK regression is equivalent to using gradient descent to train an infinite-wide neural network. The key benefit of using GNTK is that, similar to any kernel method, GNTK's parameters can be solved directly in a single step. This can avoid time-consuming gradient descent. Meanwhile, sketching has become increasingly used in speeding up various optimization problems, including solving kernel regression. Given a kernel matrix of $n$ graphs, using sketching in solving kernel regression can reduce the running time to $o(n^3)$. But unfortunately such methods usually require extensive knowledge about the kernel matrix beforehand, while in the case of GNTK we find that the construction of the kernel matrix is already $O(n^2N^4)$, assuming each graph has $N$ nodes. The kernel matrix construction time can be a major performance bottleneck when the size of graphs $N$ increases. A natural question to ask is thus whether we can speed up the kernel matrix construction to improve GNTK regression's end-to-end running time. This paper provides the first algorithm to construct the kernel matrix in $o(n^2N^3)$ running time.

%% file: intro.tex
\section{Introduction}
Graph Neural Networks (GNNs) have quickly become a popular method for machine learning on graph data. GNNs have delivered ground-breaking results in many important areas of AI, including social networking~\cite{yzw20}, bio-informatics~\cite{zl17,ywh20}, recommendation systems~\cite{yhc18}, and autonomous driving~\cite{wwm20, ysl20}. 
Given the importance of GNNs, how to train GNNs efficiently has become one of the most important problems in the AI community. 
However, efficient GNN training is challenging due to the relentless growth in the model complexity and dataset sizes, both in terms of the number of graphs in a dataset and the sizes of the graphs. 

Recently, a new direction for fast GNN training is to use Graph Neural Tangent Kernel (GNTK). \cite{dhs+19} has shown that GNTK can achieve similar accuracy as GNNs on many important learning tasks.
At that same time, GNTK regression is significantly faster than iterative stochastic gradient descent optimization because solving the parameters in GNTK is just a single-step kernel regression process. Further, GNTK can scale with GNN model sizes because the regression running time grows only linearly with the complexity of GNN models. 

Meanwhile, sketching has been increasingly used in optimization problems~\cite{cw13,nn13,mm13,bw14,swz17,alszz18,jswz21,dly21}, including linear regression, kernel regression and linear programming. Various approaches have been proposed either to sketch down the dimension of kernel matrices then solve the low-dimensional problem, or to approximate the kernel matrices by randomly constructing low-dimensional feature vectors then solve using random feature vectors directly.

It is natural to think about whether we can use sketching to improve the running time of GNTK regression. 
Given $n$ graphs where each graph has $N$ nodes, the kernel matrix has a size of $n \times n$. Given the kernel matrix is already constructed, it is well-known that using sketching can reduce the regression time from $O(n^3)$ to $o(n^3)$ under certain conditions. However, in the context of GNTK, the construction of the kernel matrix need the running time of $O(n^2N^4)$ to begin with.
This is because for each pair of graphs, we need to compute the Kronecker product of those two graphs.
There are $n^2$ pairs in total and each pair requires $N^4$ time computation.
The end-to-end GNTK regression time is sum of the time for kernel matrix construction time and the kernel regression time.
This means, the kernel matrix construction can be a major barrier for adopting sketching to speed up the end-to-end GNTK regression time, especially when we need to use GNTK for large graphs (large $N$).

This raises an important question:
\begin{center}
    {\it Can we speed up the kernel matrix construction time?}
\end{center}

This question is worthwhile due to two reasons. First, calculating the Kronecker product for each pair of graphs to construct the kernel matrix is a significant running time bottleneck for GNTK and is a major barrier for adopting sketching to speed up GNTK. Second, unlike kernel matrix construction for traditional neural networks in NTK, the construction of the kernel matrix is a complex iterative process, and it is unclear how to use sketching to reduce its complexity.

We accelerate the GNTK constructions in two steps. 1) Instead of calculating the time consuming Kronecker product for each pair of graphs directly, we consider the multiplication of a Kronecker product with a vector as a whole and accelerate it by decoupling it into two matrix multiplications of smaller dimensions. This allows us to achieve kernel matrix construction time of $O(n^2N^3)$; 2) Further, we propose a new iterative sketching procedure to reduce the running time of the matrix multiplications in the construction while providing the guarantee that the introduced error will not destruct the generalization ability of GNTK. This allows us to further improve the kernel matrix construction time to $o(n^2N^3)$.

To summarize, this paper has the following results and contributions:
\begin{itemize}
    \item We identify that the kernel matrix construction is the major running time bottleneck for GNTK regression in large graph tasks.
    \item We improve the kernel construction time in two steps: 1) We decouple the Kronecker-vector product in GNTK construction into matrix multiplications; and 2) We design an iterative sketching algorithm to further reduce the matrix multiplication time in calculating GNTK while maintaining its generalization guarantee. Overall, we are able to reduce the running time of constructing GNTK from $O(n^2N^4)$ to $o(n^2N^3)$.
    \item We rigorously quantify the impact of the error resulted from proposed sketching and show under certain assumptions that we maintain the generalization ability of the original GNTK.
\end{itemize}

%% file: related.tex
\section{Related work}\label{sec:related}

\paragraph{Sketching}
Sketching has many applications in numerical linear algebra, such as linear regression, low-rank approximation \cite{cw13,nn13,mm13,bw14,rsw16,swz17,alszz18}, distributed problems \cite{wz16,bwz16}, federated learning \cite{syz21_iclr}, reinforcement learning \cite{wzd+20}, tensor decomposition \cite{swz19_soda}, polynomial kernels \cite{swyz21}, kronecker regression \cite{dssw18,djssw19}, John ellipsoid \cite{ccly19}, clustering \cite{emz21}, cutting plane method \cite{jlsw20}, generative adversarial networks \cite{xzz18} and linear programming \cite{lsz19,y20,jswz21,sy21,dly21}.

\paragraph{Graph neural network (GNN)} Graph neural networks are machine learning methods that can operate on graph data \cite{zhang2020deep,wu2020comprehensive,chami2020machine,zhou2020graph}, which has applications in various areas including social science \cite{wu2020graph}, natural science \cite{sanchez2018graph,fout2017protein}, knowledge graphs \cite{hamaguchi2017knowledge}, 3D perception \cite{wslsbs19, sdmr20}, autonomous driving \cite{sr20, wwmk20}, and many other research areas \cite{dai2017learning}. Due to its convincing performance, GNN has become a widely applied graph learning method recently. 

\paragraph{Neural tangent kernel (NTK)} The theory of NTK has been proposed to interpret the learnablity of neural networks. Given neural network $f:\mathcal{W}\times\mathcal{D}\longrightarrow\R$ with parameter $W\in\mathcal{W}$ and input data $x\in\mathcal{D}$, the neural tangent kernel between data $x,y$ is defined to be \cite{jgh18}
\begin{align*}
    \k_{\ntk}(x,y) :=  \E_{W\sim\mathcal{N}} \left[\left\langle \frac{\partial f(W,x)}{\partial W},\frac{\partial f(W,y)}{\partial W} \right\rangle\right].
\end{align*}
Here expectation $\E$ is over random Gaussian initialization. It has been shown under the assumption of the NTK matrix being positive-definite \cite{dzps19,adh+19,adhlsw19,sy19,lsswy20,bpsw21,syz21_neurips,szz21}) or separability of training data points \cite{ll18,als19a,syz21_neurips,als19b}), training (regularized) neural network is equivalent to solving the neural tangent kernel (ridge) regression as long as the neural network is polynomial sufficiently wide. In the case of graph neural network, we study the corresponding graph neural tangent kernel \cite{dhs+19}.

%% file: problem.tex
\paragraph{Notations.}

Let $n$ be an integer, we define $[n] := \{1,2,\cdots,n\}$.
 For a full rank square matrix $A$, we use $A^{-1}$ to denote its true inverse. We define the big O notation such that $f(n) = O(g(n))$ means there exists $n_0 \in \mathbb{N}_+$ and $M \in \R$ such that $f(n) \leq M \cdot g(n)$ for all $n \geq n_0$. 
For a matrix $A$, we use $\| A \|$ or $\|A\|_2$ to denote its spectral norm. Let $\| A \|_F$  denote its Frobenius norm. Let $A^\top$  denote the transpose of $A$.  
For a matrix $A$ and a vector $x$, we define $\| x \|_{A} := \sqrt{ x^\top A x }$. 
We use $\phi$ to denote the ReLU activation function, i.e. $\phi(z) = \max\{z,0\}$.

\section{Preliminaries}
In this work, we consider a vanilla graph neural network (GNN) model consisting of three operations: {\aggregate}, {\combine} and {\readout}. In each level of GNN, we have one {\aggregate} operation followed by a {\combine} operation, which consists of $R$ fully-connected layers. At the end of the $L$ level, the GNN outputs using a {\readout} operation, which can be viewed as a pooling operation.

Specifically, let $G = (U, E)$ be a graph with node set $U$ and edge set $E$. Assume it contains $|U| = N$ nodes. Each node $u \in U$ is given a feature vector $h_u \in \R^d$. We formally define the graph neural network $f_\gnn(G)$ as follows. We use the notation $h^{(l,r)}_u $ to denote the intermediate output at level $l\in[L]$ of the GNN and layer $r\in [R]$ of the {\combine} operation.

To start with, we set the initial vector $h^{(0,R)}_u = h_u \in \R^d$, $\forall u \in U$. For the first $L$ layer, we have the following {\aggregate} and {\combine} operations.

{\bf {\aggregate} operation.} There are in total $L$ {\aggregate} operations. For any $l \in [L]$, the {\aggregate} operation aggregates the information from last level as follows:
\begin{align*}
    h^{(l, 0)}_u := c_u \cdot \sum_{a \in \neighbor(u) \cup \{u\}} h^{(l-1, R)}_a.
\end{align*}
Note that the vectors $h^{(l,0)}_u \in \R^m$ for all $l \in [2:L]$, and the only special case is $h^{(1,0)}_u \in \R^d$. $c_u \in \R$ is a scaling parameter, which controls weight of different nodes during neighborhood aggregation.

{\bf {\combine} operation.}
The {\combine} operation has $R$ fully-connected layers with ReLU activation: $\forall r \in [R]$,
\begin{align*}
    h^{(l,r)}_u := ( c_{\phi} /m )^{1/2} \cdot \phi(W^{(l, r)} \cdot h^{(l, r-1)}_u) \in \R^m.
\end{align*}
The parameters $W^{(l,r)} \in \R^{m \times m}$ for all $(l,r) \in [L] \times [R] \backslash \{(1,1)\}$, and the only special case is $W^{(1,1)} \in \R^{m \times d}$. $c_{\phi} \in \R$ is a scaling parameter, which is set to be $2$, following the initialization scheme used in \cite{dhs+19, hzr15}.

After the first $L$ layer, we have the final {\readout} operation before output.

{\bf {\readout} operation.}
We consider two different kinds of {\readout} operation.\\
1) In the simplest {\readout} operation, the final output of the GNN on graph $G$ is
\begin{align*}
    f_{\gnn}(G) := \sum_{u \in U} h^{(L, R)}_u \in \R^m.
\end{align*}
2) Using the {\readout} operation with jumping knowledge as in \cite{xltskj18}, the final output of the GNN on graph $G$ is
\begin{align*}
    f_{\gnn}(G) := \sum_{u \in U} [h^{(0, R)}_u, h^{(1, R)}_u, \cdots, h^{(L, R)}_u] \in \R^{m \times (L+1)}.
\end{align*}
When the context is clear, we also write $f_{\gnn}(G)$ as $f_{\gnn}(W, G)$, where $W$ denotes all the parameters: $W = (W^{(1,1)}, \cdots, W^{(L,R)})$.

We also introduce the following current matrix multiplication time and the notations for Kronecker product and vectorization for completeness.
\paragraph{Fast matrix multiplication.}
We use the notation $\Tmat(n,d,m)$ to denote the time of multiplying an $n \times d$ matrix with another $d \times m$ matrix. Let $\omega$ denote the exponent of matrix multiplication, i.e., $\Tmat(n,n,n) = n^{\omega}$. The first result shows $\omega < 3$ is \cite{s69}. The current best exponent is $\omega \approx 2.373$, due to \cite{w12,l14}. The common belief is $\omega \approx 2$ in the computational complexity community \cite{cksu05,w12,jswz21}. The following fact is well-known in the fast matrix multiplication literature \cite{c82,s91,bcs97} :  $\Tmat(a,b,c) = O(\Tmat(a,c,b)) = O(\Tmat(c,a,b))$ for any positive integers $a,b,c$.

\paragraph{Kronecker product and vectorization.}
Given two matrices $A \in \R^{n_1 \times d_1}$ and $B \in \R^{n_2 \times d_2}$. We use $\otimes$ to denote the Kronecker product, i.e., for $C = A \otimes B \in \R^{n_1 n_2 \times d_1 d_2}$, the $(i_1 + (i_2-1) \cdot n_1,j_1 + (j_2-1) \cdot d_1)$-th entry of $C$ is $A_{i_1,j_1} B_{i_2,j_2}$, $\forall i_1 \in [n_1], i_2 \in [n_2], j_1 \in [d_1], j_2 \in [d_2]$.  For any given matrix $H \in \R^{d_1 \times d_2}$, we use $h = \vect(H) \in \R^{d_1 d_2}$ to denote the vector such that $h_{j_1 + (j_2-1) \cdot d_1} = H_{j_1,j_2}$, $\forall j_1 \in [d_1]$, $j_2 \in [d_2]$.

\section{Graph Neural Tangent Kernel Revisited}\label{sec:our_gntk_formulation}

\begin{figure}[!t]
\begin{center}

\ifdefined\isarxiv 
\includegraphics[trim=3.5cm 8.5cm 3.5cm 8.5cm, clip=true, width=0.7\textwidth]{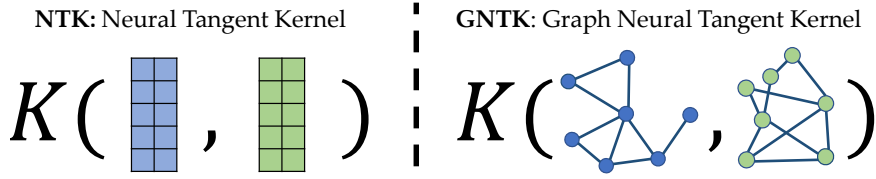}
\else 
\includegraphics[trim=3.5cm 8.5cm 3.5cm 8.5cm, clip=true, width=0.474\textwidth]{GNTK_image.pdf}
\fi
\end{center}
\ifdefined\isarxiv \else
\vspace{-0.3cm} \fi
\caption{Illustration of NTK (left) and GNTK (right). Here $\k(\cdot,\cdot)$ denotes the kernel function.} 
\label{fig:gntk}
\ifdefined\isarxiv \else
\vspace{-0.2cm} \fi
\end{figure}

In this section, we revisit the graph neural tangent kernel (GNTK) proposed in \cite{dhs+19}. A simplified illustration is shown in Fig.~\ref{fig:gntk}. Following the setting discussed in previous section, let $G = (U,E)$ and $H = (V,F)$ be two graphs with $|U| = N$ and $|V| = N'$. We use $A_G$ and $A_H$ to denote the adjacency matrix of $G$ and $H$. We give the recursive formula to compute the kernel value $\k_{\gntk}(G, H) \in \R$ induced by this GNN, which is defined as
{
\begin{align*}
     \k_{\gntk}(G, H) :=   \E_{W  \sim \N(0, I)} \Big[ \Big\langle \frac{\partial f_{\gnn}(W,G)}{\partial W}, \frac{\partial f_{\gnn}(W,H)}{\partial W} \Big\rangle \Big],
\end{align*}}
where $\N(0,I)$ is a multivariate Gaussian distribution.

Recall that the GNN uses scaling factors $c_u$ for each node $u \in G$. We define $C_G$ to be the diagonal matrix such that $(C_G)_u = c_u$ for any $u \in U$. Similarly we define $C_H$. 
For each level of {\aggregate} and {\combine} operations $\ell \in [0:L]$ and each level of fully-connected layers inside a {\combine} operation $r \in [0:R]$, we recursively define the intermediate matrices $\Sigma^{(\ell,r)}(G,H) \in \R^{N \times N'}$ and 
$K^{(\ell,r)}(G,H) \in \R^{N \times N'}$.

Initially we define $\Sigma^{(0,R)}(G,H)$, 
$K^{(0,R)}(G,H) \in \R^{N \times N'}$ as follows: $\forall u \in U, v \in V$,
\begin{align*}
     & ~ [\Sigma^{(0,R)}(G,H) ]_{u,v} := \langle h_u , h_{v} \rangle, \\
     & ~ [K^{(0,R)}(G,H) ]_{u,v} := \langle h_u, h_v \rangle.
\end{align*}
Here $h_u, h_{v} \in \R^d$ denote the input features of $u$ and $v$.
Next we recursively define $\Sigma^{(\ell,r)}(G,H)$ and 
$K^{(\ell,r)}(G,H)$ for $l \in [L]$ and $r \in [R]$ by interpreting the {\aggregate} and {\combine} operation.

\paragraph{Exact {\aggregate} operation.} 
The {\aggregate} operation gives the following formula:{
\begin{equation}\label{eq:cal_aggre}
\begin{aligned}
& ~[\Sigma^{(\ell,0)}(G,  H)]_{u, v} \\
:= & ~
c_u c_{v}  \sum_{a \in \neighbor(u) \cup \{u\}}\sum_{b \in \neighbor(v) \cup \{v\}}  [\Sigma^{(\ell-1, R)}(G,H) ]_{a, b},\\
& ~ [K^{(\ell, 0)}(G,  H) ]_{u, v} \\ 
:= & ~
c_u c_{v}  \sum_{a \in \neighbor(u) \cup \{u\}}\sum_{b \in \neighbor(v) \cup \{v\}}  [K^{(\ell-1, R)}(G,H) ]_{a , b}.
\end{aligned}
\end{equation}}

\begin{figure}[!t]
\begin{center}
\ifdefined\isarxiv 
\includegraphics[ trim=5.1cm 6cm 5.2cm 6cm, clip=true, width=0.7\textwidth]{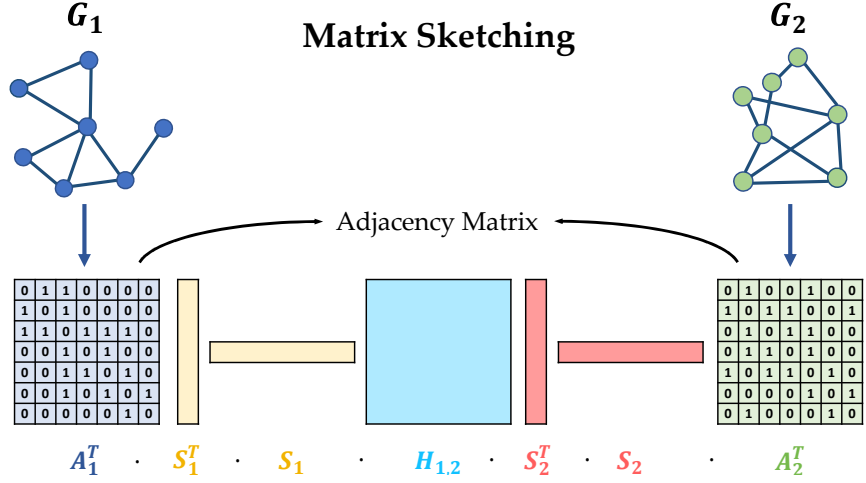}
\else 
\includegraphics[trim=5.1cm 6cm 5.2cm 6cm, clip=true, width=0.474\textwidth]{sketching_image.pdf}
\fi
\end{center}
\ifdefined\isarxiv \else
\vspace{-0.3cm} \fi
\caption{Illustration of our proposed sketching mechanism. Graphs $G_1,G_2$ are represented as adjacent matrices $A_1,A_2$. We apply sketching matrices to approximately calculate $A_1^\top H_{1,2} A_2^\top$.} 
\label{fig:sketching}
\ifdefined\isarxiv \else
\vspace{-0.2cm} \fi
\end{figure}

\paragraph{Exact {\combine} operation.} 
The {\combine} operation has $R$ fully-connected layers with ReLU activation $\phi(z) = \max\{0, z\}$.
We use $\dot{\phi}(z) = \mathbbm{1}[z \ge 0]$ to denote be the derivative of $\phi$.

For each $r \in [R]$, for each $u \in U$ and $v \in V$, we define a covariance matrix
\begin{align*}
& ~ [A^{(\ell, r)} (G,H)  ]_{u, v} \\
:= & ~\begin{pmatrix}
 [ \Sigma^{(\ell, r-1)}(G,G)  ]_{u, u}  &  [ \Sigma^{(\ell, r-1)}(G,H)  ]_{u, v} \\
 [\Sigma^{(\ell, r-1)}(H,G) ]_{u, v} &  [\Sigma^{(\ell, r-1)}(H,H) ]_{v, v} 
\end{pmatrix} \in \mathbb{R}^{2 \times 2}.
\end{align*}
Then we recursively define $[\Sigma^{(\ell, r)}(G,H)]_{u,v}$ and $[K^{(\ell, r)}(G,H)]_{u,v}$ as follows:{
\begin{align*}
 [ \Sigma^{(\ell, r)}(G, H)  ]_{u, v} := & ~ c_{\phi} \cdot \E_{(a,b) \sim \N (0, [A^{(\ell, r)} (G,H) ]_{u, v})} \big[\phi(a) \phi(b) \big], \\
 [\dot{\Sigma}^{(\ell, r)} (G, H)  ]_{u, v} := & ~ c_\phi \cdot \E_{(a,b) \sim \N(0, [A^{(\ell, r)} (G, H)]_{u, v})} \big[ \dot{\phi}(a) \dot{\phi}(b) \big],  \\
 [K^{(\ell, r)}(G, H) ]_{u, v} := & ~
 ~  [K^{(\ell, r-1)}(G,H) ]_{u, v} \cdot  [\dot{\Sigma}^{(\ell, r)}(G, H) ]_{u, v} \\
 & ~ + 
 ~[\Sigma^{(\ell, r)}(G, H) ]_{u, v}.
\end{align*}}

The intermediate results will be used to calculate the final output of the corresponding GNTK.

\paragraph{Exact {\readout} operation.}
As the final step, we compute $\k_{\gntk}(G,H) \in \R$ using the intermediate matrices. This step corresponds to the {\readout} operation.

If we do not use jumping knowledge,
\begin{align*}
	\k_{\gntk}(G,H) =
	\sum_{u \in U, v \in V}  [K^{(L, R)}(G, H) ]_{u, v}.
\end{align*}

If we use jumping knowledge,
\begin{align*}
	\k_{\gntk}(G,H) =
	\sum_{u \in U, v \in V} \sum_{l=0}^L [K^{(l, R)}(G, H) ]_{u, v}.
\end{align*}

We briefly review the running time in previous work. 
\begin{theorem}[Running time of \cite{dhs+19}, simplified version of Theorem~\ref{thm:time_dhs}]\label{thm:time_dhs_informal}
Consider a GNN with $L$ {\aggregate} operations and $L$ {\combine} operations, and each {\combine} operation has $R$ fully-connected layers. We compute the kernel matrix using $n$ graphs $\{G_i=(V_i,E_i)\}_{i=1}^n$ with $|V_i| \leq N$. Let $d \in \mathbb{N}_+$ be the dimension of the feature vectors. The total running time is
\begin{align*}
    O(n^2) \cdot (\Tmat(N, N, d) + L \cdot N^4 + LR \cdot N^2).
\end{align*}
\end{theorem}
When using GNN, we usually use constant number of operations and fully-connected layers, i.e., $L=O(1), R = O(1)$, and we have $d=o(N)$, while the size of the graphs can grow arbitrarily large. Thus it is easy to see that the dominating term in the above running time is $O(n^2N^4)$.

\section{Approximate GNTK via iterative sketching}\label{sec:approx_gntk_formula}
Now we consider the running time of solving GNTK regression. Despite rich related work to accelerate kernel regression by constructing random feature vector using sampling and sketching methods, they all require sufficient knowledge of the kernel matrix itself. While Theorem~\ref{thm:time_dhs_informal} shows the running complexity of constructing GNTK can be as large as $O(n^2N^4)$, which dominates the overall computation complexity especially in the case of relative large graphs $n = o(N^2)$. Therefore, our main task focus on a fast construction of the GNTK by randomization techniques, while ensuring the resulting GNTK regression give same generalization guarantee.

To compute an approximate version of the kernel value $\wt{K}(G, H) \in \R$ such that  $\wt{\k}_{\gntk}(G, H) \approx \k_{\gntk}(G, H)$, we will use the notation $\wt{\Sigma}^{(\ell,r)}(G,H) \in \R^{N \times N'}$ and $\wt{K}^{(\ell,r)}(G,H) \in \R^{N \times N'}$ to denote the approximated intermediate matrices for each $\ell \in [0:L]$ and $r \in [0:R]$. We observe that the computation bottleneck is to conduct the {\aggregate} operation~\eqref{eq:cal_aggre}, which takes $O(N^4)$ for obtaining ${\Sigma}^{(\ell,0)}(G,H)$ and ${K}^{(\ell,0)}(G,H)$. Hence we apply random sketching matrices $S_G \in \R^{b \times N}$ and $S_H \in \R^{b' \times N'}$ iteratively to accelerate the computation. Our proposed sketching mechanism is shown in Fig.~\ref{fig:sketching}.

\paragraph{Approximate {\aggregate} operation.}
We note that Eq.~\eqref{eq:cal_aggre} can be equivalently rewrite in following forms using Kronecker product:
\begin{equation}\label{eq:intro_exact_kron_1}
\begin{aligned}
\vect({\Sigma}^{(\ell,0)}(G,H)) := & ~ ((C_{G} A_{G}) \otimes (C_{H} A_{H})) \\
& ~\cdot \vect({\Sigma}^{(\ell-1, R)}(G,H)),\\
 \vect({K}^{(\ell,0)}(G,H)) := & ~ ( (C_{G} A_{G}) \otimes (C_{H} A_{H}) ) \\
 & ~ \cdot \vect({K}^{(\ell-1, R)}(G,H)).
\end{aligned}
\end{equation}
Now we make the following key observation about kronecker product and vectorization. Proof is delayed to Section~\ref{sec:kron_sketching_proof}.
\begin{fact}[Equivalence between two matrix products and Kronecker product then matrix-vector multiplication]\label{fac:equivalence_matrix_product_kronecker}
Given matrices $A \in \R^{n_1 \times d_1}$, $B \in \R^{n_2 \times d_2}$, and $H \in \R^{d_1 \times d_2}$, we have $\vect(A H B^\top) = (A \otimes B) \cdot \vect(H)$.
\end{fact}
Above fact implies the intermediate matrices ${\Sigma}^{(\ell,0)}(G,H)$ and ${K}^{(\ell,0)}(G,H)$ can be calculated by {
\begin{equation}\label{eq:equiva_cal}
\begin{aligned}
&~ {\Sigma}^{(\ell,0)}(G,H) :=
~ C_G A_G \cdot \wt{\Sigma}^{(\ell-1, R)}(G,H) \cdot  A_H C_H,  \\
& ~ {K}^{(\ell,0)}(G,H) :=  
~ C_G A_G \cdot \wt{K}^{(\ell-1, R)}(G,H) \cdot A_H C_H.
\end{aligned}
\end{equation}}
We emphasize that the above Eq.~\eqref{eq:equiva_cal} calculates matrix production instead of Kronecker product, which reduces the running time from $O(N^4)$ to $O(\Tmat(N,N,N))$. This is our first improvement in running time.

Further, we propose to introduce sketching matrices $S_G \in \R^{b \times N}$ and $S_H \in \R^{b' \times N'}$ iteratively into Eq.~\eqref{eq:equiva_cal} as follows:
\begin{equation}\label{eq:intro_approx_1}
\begin{aligned}
 \wt{\Sigma}^{(\ell,0)}(G,H) := & ~
~ C_G A_G \cdot \blue{(S_G^{\top} S_G)}  \cdot \wt{\Sigma}^{(\ell-1, R)}(G,H) \\
& ~  \cdot \blue{(S_H^{\top} S_H)} \cdot A_H C_H,  \\
 \wt{K}^{(\ell,0)}(G,H) := & ~  
~ C_G A_G \cdot \blue{(S_G^{\top} S_G)} \cdot  \wt{K}^{(\ell-1, R)}(G,H)  \\
& ~ \cdot  \blue{(S_H^{\top} S_H)} \cdot A_H C_H.
\end{aligned}
\end{equation}
Note that for the special case $S_G^\top S_G = S_H^\top S_H = I$, the Eq.~\eqref{eq:intro_approx_1}
degenerates to the original case. Such randomization ensures the sketched version approximates the exact matrix multiplication as in the following Lemma, which justifies our approach to speed up calculation.
\begin{lemma}[Informal, Error bound of adding sketching]\label{lem:CE_two_sketch}
Let $S_i \in \R^{b \times N}$'s denote independent AMS matrices \cite{ams99}. Then for any given $n^2$ matrices $H_{1,1} , \cdots, H_{n,n} \in \R^{N \times N}$ and $n$ matrices $A_{1}, \cdots, A_n \in \R^{N \times N}$, we have the following guarantee with high probability: for all $i , j \in [n]$,
$
    A_i^\top \blue{S_i^\top S_i} H_{i,j} \blue{S_j^\top S_j} A_j \approx A_i^\top H_{i,j} A_j.
$
\end{lemma}
Apart from above mentioned AMS sketching matrices, we point out other well-known sketching matrices such as random Gaussian, SRHT~\cite{ldfu13}, count-sketch~\cite{ccf02}, sparse embedding matrices~\cite{nn13} also apply in this case.

In the end, the {\textbf{\combine}} and {\textbf{\readout}} operation are the same as in the exact case, except now we are always working with the approximated intermediate matrices $\wt{\Sigma}^{(\ell,0)}(G,H)$ and $\wt{K}^{(\ell,0)}(G,H)$.

%% file: kronecker.tex
\subsection{Running time analysis}\label{sec:our_techniques}

The main contribution of our paper is to show that we can accelerate the computation of GNTK defined in \cite{dhs+19}, while maintaining a similar generalization bound.

In this section we first present our main running time improvement theorem.
\begin{theorem}[Main theorem, running time, Theorem~\ref{thm:time_main_formal}]\label{thm:time_main_informal}
Consider a GNN with $L$ {\aggregate} operations and $L$ {\combine} operations, and each {\combine} operation has $R$ fully-connected layers. We compute the kernel matrix using $n$ graphs $\{G_i=(V_i,E_i)\}_{i=1}^n$ with $|V_i| \leq N$. Let $b = o(N)$ be the sketch size. Let $d \in \mathbb{N}_+$ be the dimension of the feature vectors.
The total running time is
\begin{align*}
    O(n^2) \cdot (\Tmat(N, N, d) + L \cdot \Tmat(N, N, b) + LR \cdot N^2).
\end{align*}
\end{theorem}
\begin{proof}[Proof Sketch:]
Our main improvement focus on the {\aggregate} operations, where we first use the equivalent form~\eqref{eq:equiva_cal} to accelerate the exact computation from $O(N^4)$ to $\Tmat(N,N,N)$.

Further, by introducing the iterative sketching~\eqref{eq:intro_approx_1}, with an appropriate ordering of computation, we can avoid the time-consuming step of multiplying two $N \times N$ matrices. Specifically, by denoting $A_i = C_{G_i} A_{G_i}, A_j = C_{G_j} A_{G_j}, H_{i,j} = \wt{K}^{(l-1,R)}(G_i,G_j)$, we compute Eq.~\eqref{eq:intro_approx_1}, i.e., $A_i^\top S_i^\top S_i H_{i,j} S_j S_j^\top A_j$ in the following order:
\begin{itemize}
    \item  $A_i^{\top} S_i^{\top}$ and $S_j A_j$ both takes $\Tmat(N,N,b)$ time.
    \item  $S_i \cdot H_{i,j}$ takes $\Tmat(b,N,N)$ time. 
    \item  $(S_i H_{i,j}) \cdot S_j^{\top}$ takes $\Tmat(b,N,b)$ time.
    \item  $(A_i^{\top} S_i^{\top}) \cdot (S_i H_{i,j} S_j)$ takes $\Tmat(N, b, b)$ time.
    \item $(A_i^{\top} S_i^{\top} S_i H_{i,j} S_j) \cdot (S_j A_j)$ takes $\Tmat(N, b, N)$ time.
\end{itemize}
Thus, we improve the running time from $\Tmat(N,N,N)$ to $\Tmat(N,N,b)$.
\end{proof}
Note that we improve the dominating term from $N^4$ to $\Tmat(N, N, b)$.\footnote{For the detailed running time comparison between our paper and \cite{dhs+19}, please see Table~\ref{tab:my_label} in Section~\ref{sec:app_gntk_formula}.} We achieve this improvement using two techniques:
\begin{enumerate}
    \item We accelerate the multiplication of a Kronecker product with a vector by decoupling it into two matrix multiplications of smaller dimensions. In this way we improve the running time from $N^4$ down to $\Tmat(N,N,N)$.
    \item We further accelerate the two matrix multiplications by using two sketching matrices. In this way, we improve the running time from $\Tmat(N,N,N)$ to $\Tmat(N,N,b)$. 
\end{enumerate}

\subsection{Error analysis}\label{sec:error_analysis}

In this section, we prove that the introduced error due to the added sketching in calculating GNTK can be well-bounded, thus we can prove a similar generalization result. 

We first list all the notations and assumptions we used before proving the generalization bound.

\begin{definition}[Approximate GNTK with $n$ data]\label{def:gntk_main}
Let $\{(G_i, y_i)\}_{i=1}^n$ be the training data and labels, and $G_i = (V_i, E_i)$ with $|V_i| = N_i$, and we assume $N_i = O(N)$, $\forall i \in [n]$. For each $i \in [n]$ and each $u \in V_i$, let $h_u \in \R_+^d$ be the feature vector for $u$, and we define feature matrix
\begin{align*}
    H_{G_i} := [h_{u_1}, h_{u_2}, \cdots, h_{u_{N_i}}] \in \R_+^{d \times N_i}.
\end{align*}
We also define $A_{G_i} \in \R^{N_i \times N_i}$ to be the adjacency matrix of $G_i$, and $S_{G_i} \in \R^{b_i \times N_i}$ to be the sketching matrix used for $G_i$.

Let $\wt{K} \in \R^{n \times n}$ be the approximate GNTK of a GNN that has one {\aggregate} operation followed by one {\combine} operation with one fully-connected layer ($L=1$ and $R=1$) and without jumping knowledge.
For each $l \in [L]$, $r \in [R]$, $i,j \in [n]$, let $\wt{\Sigma}^{(l,r)}(G_i,G_j), \wt{K}^{(l,r)}(G_i,G_j) \in \R^{N_i \times N_j}$ be defined as in Eq.~\eqref{eq:intro_approx_1}.

We set the scaling parameters used by the GNN for $G_i$ are $c_{\phi} = 2$ and $c_u = (\|[H_{G_i} S_{G_i}^{\top} S_{G_i} A_{G_i}]_{*,u} \|_2)^{-1}$, for each $u \in V_i$. We use $C_{G_i} \in \R^{N_i \times N_i}$ to denote the diagonal matrix with $[C_{G_i}]_{u, u} = c_u$.

We further define two vectors for each $i \in [n]$ and each $u \in V_i$:
\begin{align}
    \ov{h}_{u} := &~ [H_{G_i} A_{G_i} C_{G_i}]_{*,u} \in \R^d, \label{eq:pf1_ov_h_u}\\
    \wt{h}_{u} := &~ [H_{G_i} S_{G_i}^{\top} S_{G_i} A_{G_i} C_{G_i}]_{*,u} \in \R^d. \label{eq:pf1_wt_h_u}
\end{align}
And let $T \in \mathbb{N}_+$ be a integer. For each $t \in \mathbb{N}_+$, we define two matrices $\ov{H}^{(t)}, \wt{H}^{(t)} \in \R^{d \times n}$:{
\begin{align}
    \ov{H}^{(t)} := \Big[ \sum_{u \in V_1} \Phi^{(t)}(\ov{h}_u), \cdots, \sum_{u \in V_n} \Phi^{(t)}(\ov{h}_u) \Big] \in \R^{d \times n}, \label{eq:pf1_ov_H_2l} \\
    \wt{H}^{(t)} := \Big[ \sum_{u \in V_1} \Phi^{(t)}(\wt{h}_u), \cdots, \sum_{u \in V_n} \Phi^{(t)}(\wt{h}_u) \Big] \in \R^{d \times n}, \label{eq:pf1_wt_H_2l}
\end{align}}
where we define $\Phi^{(t)}(\cdot)$ to be the feature map of the polynomial kernel of degree $t$ s.t.
\begin{align*}
\langle x, y \rangle^{t} = \langle \Phi^{(t)}(x), \Phi^{(t)}(y) \rangle ~~~ \forall x,y \in \R^d.
\end{align*}
\end{definition}

\begin{assumption}
\label{ass:gntk_main}
We assume the following properties about the input graphs, its feature vectors, and its labels.
\begin{enumerate}
    \item {\bf Labels.} 
    We assume for all $i \in [n]$, the label $y_i \in \R$ we want to learn satisfies
    \begin{align}\label{eq:learn_obj}
        y_i &= \alpha_1 \sum_{u \in V_i} \langle \ov{h}_u  , \beta_1 \rangle + \sum_{l = 1}^{T}  \alpha_{2l} \sum_{u \in V_i}  \langle \ov{h}_u , \beta_{2l} \rangle ^{2l},
    \end{align}
    Note that $\alpha_1, \alpha_2, \cdots,\alpha_{2T}$ are scalars, $\beta_1, \beta_2, \cdots,\beta_{2T} $ are vectors in $d$-dimensional space.
    \item {\bf Feature vectors and graphs.} For each $t \in \{1\} \cup \{2l\}_{l=1}^T$, we assume we have
    $
        \|(\ov{H}^{(t)})^{\top} \ov{H}^{(t)}\|_F \leq \gamma_t \cdot \|(\ov{H}^{(t)})^{\top} \ov{H}^{(t)}\|_2,
    $
    where $\gamma_1, \gamma_2, \gamma_4, \cdots, \gamma_{2T} \in \R$. We also let $\gamma = \max_{t\in \{1\} \cup \{2l\}_{l=1}^T}\{\gamma_t\}$. Note that $\gamma \geq 1$.
    \item {\bf Sketching sizes.} We assume the sketching sizes $\{b_i\}_{i=1}^n$ satisfy that $\forall i, j \in [n]$,{
    \begin{align*}
        &~ \|A_{G_j} C_{G_j} \mathbf{1}_{[N_j]}\|_2 \|A_{G_i} C_{G_i} \mathbf{1}_{[N_i]}\|_2 \cdot \|H_{G_j}^{\top} H_{G_i}\|_F \\
        \lesssim &  \frac{\min\{\sqrt{b_i}, \sqrt{b_j}\}}{\gamma T \log^3 N} 
        \mathbf{1}_{[N_i]}^{\top} C_{G_i}^{\top} A_{G_i}^{\top} H_{G_i}^{\top} H_{G_j} A_{G_j} C_{G_j} \mathbf{1}_{[N_j]},
    \end{align*}}
    where $\mathbf{1}_{[N_i]} \in \R^{N_i}, \mathbf{1}_{[N_j]} \in \R^{N_j}$ are the all one vectors of size $N_i$ and $N_j$.
\end{enumerate}
\end{assumption}

We now provide the generalization bound of our work. We start with a standard tool. 

\begin{theorem}[\cite{bm02}]\label{thm:kernel_gen_main}
Consider $n$ training data $\{(G_i, y_i)\}_{i = 1}^n$ drawn i.i.d. from distribution $\mathcal{D}$ and $1$-Lipschitz loss function $\ell : \mathbb{R} \times \mathbb{R} \to [0, 1]$ satisfying $\ell(y, y) = 0$.
Then with probability at least $1 - \delta$, the population loss of the $\mathrm{GNTK}$ predictor $f_{\gntk}$ is upper bounded by
\begin{align*}
L_{\mathcal{D}} (f_{\gntk} ) & =  \E_{(G, y) \sim \mathcal{D}} [ \ell(f_{\gntk}(G), y) ] \\
& \lesssim  ( \|y\|_{\wt{K}^{-1}}^2 \cdot \tr [ \wt{K} ] )^{1/2} / {n} 
+ \sqrt{ \log(1/\delta) / n } .
\end{align*}
\end{theorem}

Now it remains to bound $\|y\|_{\wt{K}^{-1}}$ and $\tr [ \wt{ K } ]$. 
We have the following three technical lemmas to address $\|y\|_{\wt{K}^{-1}}$ and $\tr [ \wt{ K } ]$. To start with, we give a close-form reformulation of our approximate GNTK.

\begin{lemma}[Close-form formula of approximate GNTK]\label{lem:close_form_approx_gntk_main}
Following the notations of Definition~\ref{def:gntk_main}, we can decompose $\wt{K} \in \R^{n \times n}$ into
$
    \wt{K} = \wt{K}_1 + \wt{K}_2 \succeq \wt{K}_1,
$
where $\wt{K}_2 \in \R^{n \times n}$ is a PSD matrix, and $\wt{K}_1 \in \R^{n \times n}$. 
$\wt{K}_1$ satisfies the following:
\begin{align*}
    \wt{K}_1 = \frac{1}{4} (\wt{H}^{(1)})^{\top} \cdot \wt{H}^{(1)} 
     +  \frac{1}{2\pi} \sum_{l=1}^{\infty} c_l \cdot (\wt{H}^{(2l)})^{\top} \cdot \wt{H}^{(2l)},
\end{align*}
where $c_l = \frac{(2l-3)!!}{(2l-2)!! (2l-1)}$ and equivalently for each $i,j \in [n]$, $\wt{K}_1({G_i},{G_j}) \in \R$ satisfies the following: {
\begin{align*}
     \wt{K}_1({G_i},{G_j}) 
    = \sum_{u \in V_i} \sum_{v \in V_j} \langle \wt{h}_u, \wt{h}_v \rangle \cdot \frac{1}{2\pi} \big(\pi - \arccos (\langle \wt{h}_u, \wt{h}_v \rangle) \big),
\end{align*}}
For each $i,j \in [n]$, $\wt{K}_2({G_i},{G_j}) \in \R$ satisfies the following:
\begin{align*}
     & ~\wt{K}_2({G_i},{G_j})  
    = \sum_{u \in V_i} \sum_{v \in V_j} \langle \wt{h}_u, \wt{h}_v \rangle \\
    & ~ \cdot \frac{1}{2\pi}   \Big( \pi - \arccos (\langle \wt{h}_u, \wt{h}_v \rangle) + \sqrt{1 - \langle \wt{h}_u, \wt{h}_v \rangle^2} \Big).
\end{align*}
\end{lemma}

Based upon the above characterization, we are ready to bound $y^\top \wt{K}^{-1}y$, as a generalization of Theorem~4.2 of \cite{dhs+19}.
\begin{lemma}[Bound on $y^\top \wt{K}^{-1}y$]\label{lem:learnability1_main}
Following the notations of Definition~\ref{def:gntk_main} and under the assumptions of Assumption~\ref{ass:gntk_main}, we have 
\begin{align*}
\| y \|_{ \wt{K}^{-1} } \leq 4 \cdot |\alpha_1| \|\beta_1\|_2 + \sum_{l=1}^{T} 4 \sqrt{\pi} (2l-1) \cdot |\alpha_{2l}| \|\beta_{2l}\|_2.
\end{align*}
\end{lemma}

We provide a high-level proof sketch here. We first compute all the variables in the approximate GNTK formula to get a close-form formula of $\wt{K}$. Then combining with the assumption on the labels $y$, we show that $ \|y\|_{\wt{K}_1^{-1}}$ is upper bounded by
\begin{align}\label{eq:y_k_inter}
    \|y\|_{\wt{K}_1^{-1}}
    \leq &~ (4 \alpha^2 \cdot \beta^{\top} \ov{H} \cdot (\wt{H}^{\top} \wt{H})^{-1} \cdot \ov{H}^{\top} \beta)^{1/2},
\end{align}
where $\ov{H}, \wt{H} \in \R^{d \times n}$ are two matrices such that $\forall i,j \in [n]$, 
$
     [\ov{H}^{\top} \ov{H}]_{i,j} =  \mathbf{1}_{N_i}^{\top} C_{G_i}^{\top} A_{G_i}^{\top} H_{G_i}^{\top} \cdot H_{G_j} A_{G_j} C_{G_j} \mathbf{1}_{N_j},
$
and
$
     [\wt{H}^{\top} \wt{H}]_{i,j} =  \mathbf{1}_{N_i}^{\top} C_{G_i}^{\top} A_{G_i}^{\top} \blue{(S_{G_i}^{\top} S_{G_i})} H_{G_i}^{\top}  
      \cdot  H_{G_j} \blue{(S_{G_j}^{\top} S_{G_j})} A_{G_j} C_{G_j} \mathbf{1}_{N_j}.
$
Note by Lemma~\ref{lem:CE_two_sketch}, we can show that the sketched version $\wt{H}^{\top} \wt{H}$ is a PSD approximation of $\ov{H}^{\top} \ov{H}$ in the sense of
$
(1-\frac{1}{10}) \ov{H}^{\top} \ov{H} \preceq \wt{H}^{\top} \wt{H} \preceq (1+\frac{1}{10}) \ov{H}^{\top} \ov{H}.
$
Plugging into Eq.~\ref{eq:y_k_inter} we can complete the proof.
\begin{figure}[!t]
\begin{center}
\ifdefined\isarxiv
\includegraphics[trim=0.3cm 0.4cm 0.2cm 0.4cm, clip=true, width=0.7\textwidth]{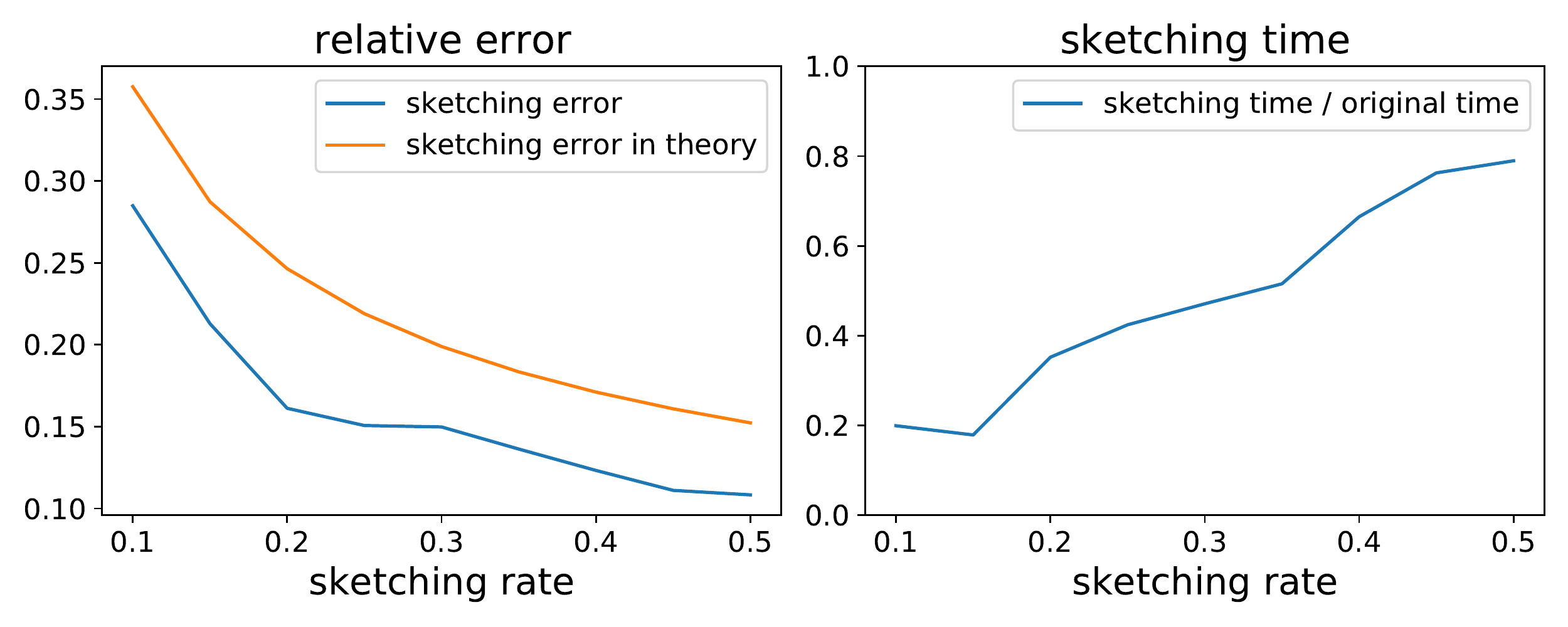}
\else 
\includegraphics[trim=0.3cm 0.4cm 0.2cm 0.4cm, clip=true, width=0.474\textwidth]{sk_error_time.pdf}
\fi
\end{center}
\ifdefined\isarxiv \else
\vspace{-0.3cm} \fi
\caption{Comparison between theoretical and experimental sketching errors (left) and sketching time (right) under different sketching rates. } 
\label{fig:sk_error_time}
\ifdefined\isarxiv \else
\vspace{-0.2cm} \fi
\end{figure}

Lastly, we give a bound on the trace of $\wt{K}$. 
We defer the proof to Appendix.
\begin{lemma}[Bound on trace of $\wt{K}$]\label{lem:trace_main}
Following the notations of Definition~\ref{def:gntk_main} and under the assumptions of Assumption~\ref{ass:gntk_main}, we have
$
    \tr[\wt{K}] \leq 2 n N^2.
$
\end{lemma}

Combining Lemma~\ref{lem:learnability1_main} and Lemma~\ref{lem:trace_main} with Theorem~\ref{thm:kernel_gen_main}, we conclude with the following main generalization theorem:
\begin{theorem}[Main generalization theorem]\label{thm:main_generalization_main}
Let $c \in (0,1)$ denote a fixed constant. Following the notations of Definition~\ref{def:gntk_main}, and under the assumptions of Assumption~\ref{ass:gntk_main},
if we further have the conditions that
\begin{align*}
4 \cdot \alpha_1 \|\beta_1\|_2 + \sum_{l=1}^{T} 4 \sqrt{\pi} (2l-1) \cdot \alpha_{2l} \|\beta_{2l}\|_2 &= o(n) \end{align*}
and $N = o(\sqrt{n})$, 
then we can upper bound the generalization error of the approximate GNTK by 
\begin{align*}
L_{\mathcal{D}} (f_{\gntk} ) =  \E_{(G, y) \sim \mathcal{D}} [ \ell(f_{\gntk}(G), y) ] \lesssim  O(1/n^c).
\end{align*}
\end{theorem}
Above theorem shows that the approximate GNTK corresponds to the vanilla GNN described above is able to, with polynomial number of samples, learn functions of forms in \eqref{eq:learn_obj}. Such a guarantee is similar to the result for exact GNTK~\cite{adh+19}, indicating our proposed sketching does not influence the generalization ability of GNTK.

We conduct experiments to validate that the error introduced by matrix sketching is strictly bounded. Following Lemma~\ref{lem:CE_two_sketch} , we validate the error difference between matrix multiplication with and without the sketching method. Specifically, we randomly generate $n \times n$  matrices $A$, $G$ and $H$. And matrix multiplication without sketching is calculated by $M=G^\top A H$. For the sketching method, we randomly generate two AMS matrices $R$ and $S$ with size $\gamma n \times n$ where $\gamma$ is the sketching ratio. And matrix multiplication with sketching is calculated by $M_{\text{sketch}}=G^\top R^\top R A S^\top S H$. The experimental error matrix is calculated by $|M-M_{\text{sketch}}|$, and the theoretical error matrix is calculated by the RHS of Lemma~\ref{lem:CE_two_sketch}. We divide both errors by the original matrix $M$ to show the relative error. And we show the final mean error by taking the average over all entries of the error matrices. 

Fig. \ref{fig:sk_error_time} shows the result. We set $n=500$, and run experiments under different sketching rates from $0.1$ to $0.9$. We run each sketching rate for $100$ times and calculate the mean error. We also show the matrix multiplication time with/without sketching. Experiments show that our sketching error is always lower than the theoretical bound. When the sketching rate is high, the error decreases and the running time increases because the dimension of the matrix is larger. This experiment validates our Lemma~\ref{lem:CE_two_sketch}, showing that our matrix sketching method has a strictly bounded error.

%% file: conclusion.tex
\section{ Conclusion}\label{sec:discuss}

Graph Neural Networks (GNNs) have recently become the most important method for machine learning on graph data (e.g., protein structures, code AST, social networks), but training GNNs efficiently is a major challenge. An alternative method is Graph Neural Tangent Kernel (GNTK). 
GNTK's parameters are solved directly in a single step. This avoids time-consuming gradient descent. GNTK has thus become the state-of-the-art method to achieve high training speed without compromising accuracy.
In this paper, we accelerate the construction of GNTK by two steps: 1) accelerate the multiplication of a Kronecker product with a vector by decoupling it into two matrix multiplications of smaller dimensions; 2) we introduce sketching matrices iteratively to further accelerate the multiplication between two matrices. 
Our techniques speed up generating kernel matrices for GNTK and thus improve the end-to-end running time for GNTK regression.

%% file: app_preli.tex
\section*{Appendix}

In Section~\ref{sec:app_preli}, we define several basic notations. In Section~\ref{sec:app_gntk_formula}, we review the GNTK definitions of \cite{dhs+19} and then define our approximate version of GNTK formulas. In Section~\ref{sec:generalization_bound}, we prove the generalization bound for one layer approximate GNTK (previously presented in Theorem~\ref{thm:main_generalization_main}). In Section~\ref{sec:time}, we formally analyze the running time to compute our approximate GNTK. In Section~\ref{sec:kron_sketching_proof}, we prove some tools of Kronecker product and sketching. In Section~\ref{sec:exp}, we conduct some experiments on our matrix decoupling method to validate our results.

\section{Preliminaries}\label{sec:app_preli}

{\bf Standard notations.} 
For a positive integer $n$, we define $[n] := \{1,2,\cdots,n\}$.
 For a full rank square matrix $A$, we use $A^{-1}$ to denote its true inverse. We define the big O notation such that $f(n) = O(g(n))$ means there exists $n_0 \in \mathbb{N}_+$ and $M \in \R$ such that $f(n) \leq M \cdot g(n)$ for all $n \geq n_0$.

\noindent{\bf Norms.}
For a matrix $A$, we use $\| A \|$ or $\|A\|_2$ to denote its spectral norm. We use $\| A \|_F$ to denote its Frobenius norm. We use $A^\top$ to denote the transpose of $A$.  
For a matrix $A$ and a vector $x$, we define $\| x \|_{A} := \sqrt{ x^\top A x }$.

\noindent{\bf Functions.} 
We use $\phi$ to denote the ReLU activation function, i.e. $\phi(z) = \max\{z,0\}$. 
For a function $f:\R \to \R$, we use $f'$ to denote the derivative of $f$.

\paragraph{Fast matrix multiplication.}
We define the notation $\Tmat(n,d,m)$ to denote the time of multiplying an $n \times d$ matrix with another $d \times m$ matrix. Let $\omega$ denote the exponent of matrix multiplication, i.e., $\Tmat(n,n,n) = n^{\omega}$. The first result shows $\omega < 3$ is \cite{s69}. The current best $\omega \approx 2.373$ due to \cite{w12,l14}. The following fact is well-known in the fast matrix multiplication literature \cite{c82,s91,bcs97} :  $\Tmat(a,b,c) = O(\Tmat(a,c,b)) = O(\Tmat(c,a,b))$ for any positive integers $a,b,c$.

\paragraph{Kronecker product and vectorization.}
Given two matrices $A \in \R^{n_1 \times d_1}$ and $B \in \R^{n_2 \times d_2}$. We use $\otimes$ to denote the Kronecker product, i.e., for $C = A \otimes B \in \R^{n_1 n_2 \times d_1 d_2}$, the $(i_1 + (i_2-1) \cdot n_1,j_1 + (j_2-1) \cdot d_1)$-th entry of $C$ is $A_{i_1,j_1} B_{i_2,j_2}$, $\forall i_1 \in [n_1], i_2 \in [n_2], j_1 \in [d_1], j_2 \in [d_2]$.  For any give matrix $H \in \R^{d_1 \times d_2}$, we use $h = \vect(H) \in \R^{d_1 d_2}$ to denote the vector such that $h_{j_1 + (j_2-1) \cdot d_1} = H_{j_1,j_2}$, $\forall j_1 \in [d_1]$, $j_2 \in [d_2]$.

\input{probability}

%% file: gntk_formula.tex
\section{GNTK formulas}\label{sec:app_gntk_formula}
In this section, we first present the GNTK formulas for GNNs of \cite{dhs+19}, we then show our approximate version of the GNTK formulas.

\subsection{GNNs}\label{sec:gnn}
A GNN has $L$ {\aggregate} operations, each followed by a {\combine} operation, and each {\combine} operation has $R$ fully-connected layers. The fully-connected layers have output dimension $m$, and use ReLU as non-linearity. In the end the GNN has a {\readout} operation that corresponds to the pooling operation of normal neural networks.

\begin{table*}[h]
    \centering
    \begin{tabular}{|l|l|} \hline
        {\bf Reference} & {\bf Time} \\ \hline
        \cite{dhs+19} & $O(n^2) \cdot (\Tmat(N, N, d) + L \cdot N^4 + LR \cdot N^2)$ \\ \hline
         Thm.~\ref{thm:time_main_informal} and \ref{thm:main_generalization_main} & $O(n^2) \cdot (\Tmat(N, N, d) + L \cdot \Tmat(N, N, b) + LR \cdot N^2)$ \\ \hline
    \end{tabular}
    \caption{ When $L= O(1)$ and $R=O(1)$, the dominate term in previous work is $O(N^4)$. We improve it to $\Tmat(N,N,b)$.}
    \label{tab:my_label}
\end{table*}

Let $G = (U, E)$ be a graph with $|U| = N$ number of nodes. Each node $u \in U$ has a feature vector $h_u \in \R^d$.

We define the initial vector $h^{(0,R)}_u = h_u \in \R^d$, $\forall u \in U$.

{\bf {\aggregate} operation.} There are in total $L$ {\aggregate} operations. For any $l \in [L]$, the {\aggregate} operation aggregates the information from last level as follows:
\begin{align*}
    h^{(l, 0)}_u := c_u \cdot \sum_{a \in \neighbor(u) \cup \{u\}} h^{(l-1, R)}_a.
\end{align*}

Note that the vectors $h^{(l,0)}_u \in \R^m$ for all $l \in [2:L]$, and the only special case is $h^{(1,0)}_u \in \R^d$. $c_u \in \R$ is a scaling parameter, which controls weight of different nodes during neighborhood aggregation. In our experiments we choose $c_u$ between values $ \{1, \frac{1}{|\mathcal{N}(u)|+1} \}$ following \cite{dhs+19}.

{\bf {\combine} operation.}
The {\combine} operation has $R$ fully-connected layers with ReLU activation: $\forall r \in [R]$,
\begin{align*}
    h^{(l,r)}_u := ( c_{\phi} /m )^{1/2} \cdot \phi(W^{(l, r)} \cdot h^{(l, r-1)}_u) \in \R^m.
\end{align*}
The parameters $W^{(l,r)} \in \R^{m \times m}$ for all $(l,r) \in [L] \times [R] \backslash \{(1,1)\}$, and the only special case is $W^{(1,1)} \in \R^{m \times d}$. $c_{\phi} \in \R$ is a scaling parameter, in our experiments we set $c_{\phi}$ to be $2$, following the initialization scheme used in \cite{dhs+19, hzr15}.

{\bf {\readout} operation.}
Using the simplest {\readout} operation, the final output of the GNN on graph $G$ is
\begin{align*}
    f_{\gnn}(G) := \sum_{u \in U} h^{(L, R)}_u \in \R^m.
\end{align*}

Using the {\readout} operation with jumping knowledge as in \cite{xltskj18}, the final output of the GNN on graph $G$ is
\begin{align*}
    f_{\gnn}(G) := \sum_{u \in U} [h^{(0, R)}_u, h^{(1, R)}_u, \cdots, h^{(L, R)}_u] \in \R^{m \times (L+1)}.
\end{align*}

When the context is clear, we also write $f_{\gnn}(G)$ as $f_{\gnn}(W, G)$, where $W$ denotes all the parameters: $W = (W^{(1,1)}, \cdots, W^{(L,R)})$.

\subsection{Exact GNTK formulas}\label{sec:app_exact_gntk_formula}
We first present the GNTK formulas of \cite{dhs+19}.

We consider a GNN with $L$ {\aggregate} operations and $L$ {\combine} operations, and each {\combine} operation has $R$ fully-connected layers. We use $h(W,G) \in \R^m$ to denote the function corresponding to this GNN, where $W := \cup_{\ell \in [L], r \in [R]} \{W^{(\ell,r)}\}$ denotes all tunable parameters of this GNN. 

Let $G = (U,E)$ and $H = (V,F)$ be two graphs with $|U| = N$ and $|V| = N'$. We use $A_G \in \R^{N \times N}$ and $A_H \in \R^{N' \times N'}$ to denote the adjacency matrix of $G$ and $H$. We give the recursive formula to compute the kernel value $\k_{\gntk}(G, H) \in \R$ induced by this GNN, which is defined as
\begin{align*}
    \k_{\gntk}(G, H) := \E_{W  \sim \N(0, I)} \Big[\lim_{m \to \infty} \Big\langle \frac{\partial f_{\gnn}(W,G)}{\partial W}, \frac{\partial f_{\gnn}(W,H)}{\partial W} \Big\rangle \Big].
\end{align*}

Recall that the GNN uses scaling factors $c_u$ for each node $u \in G$. We define $C_G \in \R^{N \times N}$ to be the diagonal matrix such that $(C_G)_u = c_u$ for any $u \in U$. Similarly we define $C_H \in \R^{N' \times N'}$.

We will use intermediate matrices $\Sigma^{(\ell,r)}(G,H) \in \R^{N \times N'}$ and $K^{(\ell,r)}(G,H) \in \R^{N \times N'}$ for each $\ell \in [0:L]$ and $r \in [0:R]$. 

Initially we define $\Sigma^{(0,R)}(G,H) \in \R^{N \times N'}$ as follows: $\forall u \in U, v \in V$,
\begin{align*}
     [\Sigma^{(0,R)}(G,H) ]_{u,v} := \langle h_u , h_{v} \rangle,
\end{align*}
where $h_u, h_{v} \in \R^d$ dentoes the input features of $u$ and $v$.
And we we define $K^{(0,R)}(G,H) \in \R^{N \times N'}$ as follows: $\forall u \in U, v \in V$,
\begin{align*}
     [K^{(0,R)}(G,H) ]_{u,v} := \langle h_u, h_v \rangle.
\end{align*}

Next we recursively define $\Sigma^{(\ell,r)}(G,H)$ and $K^{(\ell,r)}(G,H)$ for $l \in [L]$ and $r \in [R]$, where $l$ denotes the level of {\aggregate} and {\combine} operations, and $r$ denotes the level of fully-connected layers inside a {\combine} operation. Then we define the final output after {\readout} operation.

\paragraph{{\aggregate} operation.} 
The {\aggregate} operation gives the following formula:
\begin{align*}
 [\Sigma^{(\ell,0)}(G,H) ]_{u, v} := &~ c_u c_{v}\sum_{a \in \neighbor(u) \cup \{u\}}\sum_{b \in \neighbor(v) \cup \{v\}}  [\Sigma^{(\ell-1, R)}(G,H) ]_{a, b},\\
 [K^{(\ell, 0)}(G,H) ]_{u, v} := &~ c_u c_{v}\sum_{a \in \neighbor(u) \cup \{u\}}\sum_{b \in \neighbor(v) \cup \{v\}}  [K^{(\ell-1, R)}(G,H) ]_{a , b}.
\end{align*}
Note that the above two equations are equivalent to the following two equations:
\begin{align*}
\Sigma^{(\ell,0)}(G,H) = &~ C_G A_G \cdot \Sigma^{(\ell-1, R)}(G,H) \cdot A_H C_H, \\
K^{(\ell,0)}(G,H) = &~ C_G A_G \cdot K^{(\ell-1, R)}(G,H) \cdot A_H C_H.
\end{align*}

\paragraph{{\combine} operation.} 
The {\combine} operation has $R$ fully-connected layers with ReLU activation $\phi(z) = \max\{0, z\}$.
We use $\dot{\phi}(z) = \mathbbm{1}[z \ge 0]$ to denote be the derivative of $\phi$.

For each $r \in [R]$, for each $u \in U$ and $v \in V$, we define a covariance matrix
\begin{align*}
 [A^{(\ell, r)} (G,H)  ]_{u, v} := \begin{pmatrix}
 [ \Sigma^{(\ell, r-1)}(G,G)  ]_{u, u}  &  [ \Sigma^{(\ell, r-1)}(G,H)  ]_{u, v} \\
 [\Sigma^{(\ell, r-1)}(H,G) ]_{u, v} &  [\Sigma^{(\ell, r-1)}(H,H) ]_{v, v} 
\end{pmatrix} \in \mathbb{R}^{2 \times 2}.
\end{align*}

Then we recursively define $[\Sigma^{(\ell, r)}(G,H)]_{u,v}$ and $[K^{(\ell, r)}(G,H)]_{u,v}$ as follows:
\begin{align}
 [ \Sigma^{(\ell, r)}(G, H)  ]_{u, v} := &~ c_{\phi} \cdot \E_{(a,b) \sim \N (0, [A^{(\ell, r)} (G,H) ]_{u, v})} \big[\phi(a) \phi(b) \big], \label{equ:exp_1} \\
 [\dot{\Sigma}^{(\ell, r)} (G, H)  ]_{u, v} := &~ c_\phi \cdot \E_{(a,b) \sim \N(0, [A^{(\ell, r)} (G, H)]_{u, v})} \big[ \dot{\phi}(a) \dot{\phi}(b) \big], \label{equ:exp_2} \\
 [K^{(\ell, r)}(G, H) ]_{u, v} := &~  [K^{(\ell, r-1)}(G,H) ]_{u, v} \cdot  [\dot{\Sigma}^{(\ell, r)}(G, H) ]_{u, v} +  [\Sigma^{(\ell, r)}(G, H) ]_{u, v}. \notag
\end{align}


\paragraph{{\readout} operation.}
Finally we compute $\k_{\gntk}(G,H) \in \R$ using the intermediate matrices. This step corresponds to the {\readout} operation.

If we do not use jumping knowledge,
\begin{align*}
	\k_{\gntk}(G,H) =
	\sum_{u \in U, v \in V}  [K^{(L, R)}(G, H) ]_{u, v}.
\end{align*}

If we use jumping knowledge,
\begin{align*}
	\k_{\gntk}(G,H) =
	\sum_{u \in U, v \in V} \sum_{l=0}^L [K^{(l, R)}(G, H) ]_{u, v}.
\end{align*}

\subsection{Approximate GNTK formulas}\label{sec:app_approx_gntk_formula}
We follow the notations of previous section. Again we consider two graphs $G = (U, E)$ and $H = (V, F)$ with $|U| = N$ and $|V| = N'$.

Now the goal is to compute an approximate version of the kernel value $\wt{K}(G, H) \in \R$ such that 
\begin{align*}
\wt{\k}_{\gntk}(G, H) \approx \k_{\gntk}(G, H).
\end{align*}
We will use intermediate matrices $\wt{\Sigma}^{(\ell,r)}(G,H) \in \R^{N \times N'}$ and $\wt{K}^{(\ell,r)}(G,H) \in \R^{N \times N'}$ for each $\ell \in [0:L]$ and $r \in [0:R]$. In the approximate version we add two random Gaussian matrices $S_G \in \R^{b \times N}$ and $S_H \in \R^{b' \times N'}$ where $b \leq N$ and $b' \leq N'$ are two parameters.

Initially, $\forall u \in U, v \in V$, we define $[\wt{\Sigma}^{(0,R)}(G,H)]_{u,v} := \langle h_u, h_{v} \rangle$ and $[\wt{K}^{(0,R)}(G,H)]_{u,v} := \langle h_u, h_{v} \rangle$, same as in the exact case.

Next we recursively define $\wt{\Sigma}^{(\ell,r)}(G,H)$ and $\wt{K}^{(\ell,r)}(G,H)$ for $\ell \in [L]$ and $r \in [R]$.

\paragraph{{\aggregate} operation.}
In the approximate version, we add two sketching matrices $S_G \in \R^{b \times N}$ and $S_H \in \R^{b' \times N'}$:
\begin{align*}
\wt{\Sigma}^{(\ell,0)}(G,H) := &~ C_G A_G \cdot (S_G^{\top} S_G) \cdot \wt{\Sigma}^{(\ell-1, R)}(G,H) \cdot (S_H^{\top} S_H) \cdot A_H C_H, \\
\wt{K}^{(\ell,0)}(G,H) := &~ C_G A_G \cdot (S_G^{\top} S_G) \cdot \wt{K}^{(\ell-1, R)}(G,H) \cdot (S_H^{\top} S_H) \cdot A_H C_H,
\end{align*}
where we define $C_G \in \R^{N \times N}$ to be the diagonal matrix such that $(C_G)_u = c_u$ for any $u \in V$. Similarly we define $C_H \in \R^{N' \times N'}$.

\paragraph{{\combine} operation.}
The {\combine} operation has $R$ fully-connected layers with ReLU activation. The recursive definitions of $[\wt{A}^{(\ell, r)}(G,H)]_{u,v} \in \R^{2 \times 2}$, $[\wt{\Sigma}^{(\ell, r)}(G,H)]_{u,v} \in \R$, $[ \wt{\Sigma}_{\cdot}^{(\ell, r)}(G,H)]_{u,v} \in \R$ and $[\wt{K}^{(\ell, r)}(G,H)]_{u,v} \in \R$ are the same as in the exact case, except now we are always working with the tilde version.

\paragraph{{\readout} operation.}
Finally we compute $\wt{\k}_{\gntk}(G,H) \in \R$ using the intermediate matrices. It is also the same as in the exact case, except now we are always working with the tilde version.

%% file: gntk_generalization_bound.tex
\section{Generalization bound of approximate GNTK}\label{sec:generalization_bound}
In this section, we give the full proof of the three lemmas used to prove Theorem~\ref{thm:main_generalization_main}, the main generalization bound for the approximate version of GNTK. This generalizes the generalization bound of \cite{dhs+19}.

\subsection{Close-form formula of approximate GNTK}
\begin{lemma}[Restatement of Lemma~\ref{lem:close_form_approx_gntk_main}]\label{lem:close_form_approx_gntk}
Following the notations of Definition~\ref{def:gntk_main}, we can decompose $\wt{K} \in \R^{n \times n}$ into
\begin{align*}
    \wt{K} = \wt{K}_1 + \wt{K}_2 \succeq \wt{K}_1,
\end{align*}
where $\wt{K}_2 \in \R^{n \times n}$ is a PSD matrix, and $\wt{K}_1 \in \R^{n \times n}$. 
$\wt{K}_1$ satisfies the following:
\begin{align*}
    \wt{K}_1 = &~ \frac{1}{4} (\wt{H}^{(1)})^{\top} \cdot \wt{H}^{(1)} + \frac{1}{2\pi} \sum_{l=1}^{\infty} \frac{(2l-3)!!}{(2l-2)!! (2l-1)} \cdot (\wt{H}^{(2l)})^{\top} \cdot \wt{H}^{(2l)},
\end{align*}
and equivalently for each $i,j \in [n]$, $\wt{K}_1({G_i},{G_j}) \in \R$ satisfies the following:
\begin{align*}
    \wt{K}_1({G_i},{G_j}) = &~ \sum_{u \in V_i} \sum_{v \in V_j} \langle \wt{h}_u, \wt{h}_v \rangle \cdot \frac{1}{2\pi} \big(\pi - \arccos (\langle \wt{h}_u, \wt{h}_v \rangle) \big),
\end{align*}
For each $i,j \in [n]$, $\wt{K}_2({G_i},{G_j}) \in \R$ satisfies the following:
\begin{align*}
    \wt{K}_2({G_i},{G_j}) = &~ \sum_{u \in V_i} \sum_{v \in V_j} \langle \wt{h}_u, \wt{h}_v \rangle \cdot \frac{1}{2\pi} \Big(\pi - \arccos (\langle \wt{h}_u, \wt{h}_v \rangle) + \sqrt{1 - \langle \wt{h}_u, \wt{h}_v \rangle^2} \Big).
\end{align*}
\end{lemma}

\begin{proof}
For $i,j \in [n]$, consider the two graphs $G_i = (V_i, E_i)$ and $G_j = (V_j, E_j)$ with $|V_i| = N_i$ and $|V_j| = N_j$, we first compute the approximate GNTK formulas that corresponds to the simple GNN (with $L=1$ and $R=1$) by following the recursive formula of Section~\ref{sec:approx_gntk_formula}.

{\bf Compute $l=0$, $r=1$ variables.}
We first compute the initial variables \begin{align*}
\wt{\Sigma}^{(0,1)}(G_i, {G_j}), \wt{K}^{(0,1)}(G_i, {G_j}) \in \R^{N_i \times N_j},
\end{align*}
for any $u \in V_i$ and $v \in V_j$, we have
\begin{align}\label{eq:pf1_Sigma_0_1}
    [\wt{\Sigma}^{(0,1)}(G_i, {G_j})]_{u,v} = \langle h_u, h_{v} \rangle, ~~
    [\wt{K}^{(0,1)}(G_i, {G_j})]_{u,v} = \langle h_u, h_{v} \rangle,
\end{align}
which follows from Section~\ref{sec:approx_gntk_formula}.

{\bf Compute $l=1$, $r=0$ variables.}
We compute $\wt{\Sigma}^{(1,0)}(G_i, {G_j}) \in \R^{N_i \times N_j}$ as follows:
\begin{align}\label{eq:pf1_Sigma_1_0}
    [\wt{\Sigma}^{(1,0)}(G_i, {G_j})]_{u,v} 
    = &~ [C_{G_i} A_{G_i} \cdot (S_{G_i}^{\top} S_{G_i}) \cdot \wt{\Sigma}^{(0, 1)}({G_i},{G_j}) \cdot (S_{G_j}^{\top} S_{G_j}) \cdot A_{G_j} C_{G_j}]_{u,v} \notag \\
    = &~ [C_{G_i} A_{G_i} \cdot (S_{G_i}^{\top} S_{G_i}) \cdot H_{G_i}^{\top} H_{G_j} \cdot (S_{G_j}^{\top} S_{G_j}) \cdot A_{G_j} C_{G_j}]_{u,v} \notag \\
    = &~ \langle \wt{h}_u, \wt{h}_v \rangle,
\end{align}
where the second step uses Eq.\eqref{eq:pf1_Sigma_0_1} and the definition $H_{G_i} := [h_{u_1}, h_{u_2}, \cdots, h_{u_{N_i}}] \in \R^{d \times N_i}$ in lemma statement, and the third step follows from the definition of $\wt{h}_u, \wt{h}_v \in \R^d$ in Eq.~\eqref{eq:pf1_wt_h_u}. 
Note that we have $\|\wt{h}_u\|_2 = 1$ since $C_{G_i}$ is a diagonal matrix with $[C_{G_i}]_{u,u} = \|[H_{G_i} S_{G_i}^{\top} S_{G_i} A_{G_i} C_{G_i}]_{*, u}\|_2^{-1}$.

Then we compute $\wt{K}^{(1,0)}({G_i},{G_j}) \in \R^{N_i \times N_j}$:
\begin{align}\label{eq:pf1_Theta_1_0}
    [\wt{K}^{(1,0)}({G_i},{G_j})]_{u,v} = &~ [C_{G_i} A_{G_i} \cdot (S_{G_i}^{\top} S_{G_i}) \cdot \wt{K}^{(0, 1)}({G_i},{G_j}) \cdot (S_{G_j}^{\top} S_{G_j}) \cdot A_{G_j} C_{G_j}]_{u,v} \notag \\
    = &~ \wt{\Sigma}^{(1,0)}({G_i},{G_j}) = \langle \wt{h}_u, \wt{h}_v \rangle,
\end{align}
where the second step uses the fact that $\wt{K}^{(0,1)}({G_i},{G_j}) = \wt{\Sigma}^{(0,1)}({G_i},{G_j})$ (see Eq.~\eqref{eq:pf1_Sigma_0_1}).

{\bf Compute $l=1$, $r=1$ variables.} Next we compute $[\wt{A}^{(1, 1)}({G_i},{G_j})]_{u,v} \in \R^{2 \times 2}$ for any $u \in V_i$ and $v \in V_j$, we have
\begin{align}\label{eq:pf1_A_1_1}
    [\wt{A}^{(1, 1)}({G_i},{G_j})]_{u,v} =
    \begin{pmatrix}
    [\wt{\Sigma}^{(1,0)}({G_i},{G_i})]_{u,u} & [\wt{\Sigma}^{(1,0)}({G_i},{G_j})]_{u,v} \\
    [\wt{\Sigma}^{(1,0)}({G_j},{G_i})]_{v,u} & [\wt{\Sigma}^{(1,0)}({G_j},{G_j})]_{v,v}
    \end{pmatrix}
    = \begin{pmatrix}
    1 & \langle \wt{h}_u, \wt{h}_v \rangle \\
    \langle \wt{h}_v, \wt{h}_u \rangle & 1
    \end{pmatrix},
\end{align}
where the second step follows from Eq.~\eqref{eq:pf1_Sigma_1_0}.

Then we compute $\wt{\Sigma}_{\cdot}^{(1, 1)}({G_i},{G_j}) \in \R^{N_i \times N_j}$. For any $u \in V_i$ and $v \in V_j$, 
\begin{align}\label{eq:pf1_Sigma_dot_1_1}
    [ {\wt{\Sigma}}_{\cdot}^{(1, 1)}({G_i},{G_j})]_{u,v} = &~ c_{\phi} \cdot \E_{(a,b) \sim \N (0, [\wt{A}^{(1,1)} ({G_i},{G_j} ) ]_{u,v})} [\dot{\phi}(a) \cdot \dot{\phi}(b) ] \notag \\
    = &~ \frac{1}{2\pi} \Big(\pi - \arccos (\langle \wt{h}_u, \wt{h}_v \rangle) \Big),
\end{align}
where the second step follows from Eq.~\eqref{eq:pf1_A_1_1} and $c_{\phi} = 2$ (Definition~\ref{def:gntk_main}).  

And similarly we compute $[\wt{\Sigma}^{(1, 1)}({G_i},{G_j})]_{u,v} \in \R^{N_i \times N_j}$. For any $u \in V_i$ and $v \in V_j$,
\begin{align}\label{eq:pf1_Sigma_1_1}
    [\wt{\Sigma}^{(1, 1)}({G_i},{G_j})]_{u,v} = &~ c_{\phi} \cdot \E_{(a,b) \sim \N (0, [\wt{A}^{(1,1)} ({G_i},{G_j} ) ]_{u,v})} [\phi(a) \cdot \phi(b) ] \notag \\
    = &~ \frac{1}{2\pi} \Big(\pi - \arccos (\langle \wt{h}_u, \wt{h}_v \rangle) + \sqrt{1 - \langle \wt{h}_u, \wt{h}_v \rangle^2} \Big),
\end{align}
where the second step follows from Eq.~\eqref{eq:pf1_A_1_1} and $c_{\phi} = 2$ (Definition~\ref{def:gntk_main}).

Then we compute $\wt{K}^{(1,1)}({G_i},{G_j}) \in \R^{N_i \times N_j}$. We decompose $\wt{K}^{(1,1)}({G_i},{G_j})$ as follows:
\begin{align}\label{eq:pf1_Theta_1_1}
    \wt{K}^{(1,1)}({G_i},{G_j}) = \wt{K}^{(1,1)}_1({G_i},{G_j}) + \wt{\Sigma}^{(1,1)}({G_i}, {G_j}),
\end{align}
where we define
\begin{align}\label{eq:pf1_Theta_1_1_1_def}
    [\wt{K}^{(1,1)}_1({G_i},{G_j})]_{u,v} := &~ [\wt{K}^{(1,0)}({G_i}, {G_j}) ]_{u,v} \cdot [ {\wt{\Sigma}}_{\cdot}^{(1,1)}({G_i}, {G_j}) ]_{u,v}, ~~ \forall u \in V_i, v \in V_j.
\end{align}
The above equation follows from the definition of $\wt{K}^{(1,1)}({G_i},{G_j})$ (see Section~\ref{sec:approx_gntk_formula}). 

We then have
\begin{align}\label{eq:pf1_Theta_1_1_1_formula}
    [\wt{K}^{(1,1)}_1({G_i},{G_j})]_{u,v} = &~ \langle \wt{h}_u, \wt{h}_v \rangle \cdot \frac{1}{2\pi} \big(\pi - \arccos (\langle \wt{h}_u, \wt{h}_v \rangle) \big) \notag \\
    = &~ \langle \wt{h}_u, \wt{h}_v \rangle \cdot \frac{1}{2\pi} \Big(\pi - \big( \frac{\pi}{2} - \sum_{l=0}^{\infty} \frac{(2l-1)!!}{(2l)!!} \cdot \frac{\langle \wt{h}_u, \wt{h}_v \rangle^{2l+1}}{2l+1} \big) \Big) \notag \\ 
    = &~ \frac{1}{4} \langle \wt{h}_u, \wt{h}_v \rangle + \frac{1}{2\pi} \sum_{l=1}^{\infty} \frac{(2l-3)!!}{(2l-2)!! (2l-1)} \cdot \langle \wt{h}_u, \wt{h}_v \rangle^{2l} \notag \\
    = &~ \frac{1}{4} \langle \wt{h}_u, \wt{h}_v \rangle + \frac{1}{2\pi} \sum_{l=1}^{\infty} \frac{(2l-3)!!}{(2l-2)!! (2l-1)} \cdot \langle \Phi^{(2l)}(\wt{h}_u), \Phi^{(2l)}(\wt{h}_v) \rangle,
\end{align}
where the first step plugs Eq.~\eqref{eq:pf1_Theta_1_0} and \eqref{eq:pf1_Sigma_dot_1_1} into Eq.~\eqref{eq:pf1_Theta_1_1_1_def}, the second step uses the Taylor expansion of $\arccos(x) = \frac{\pi}{2} - \sum_{l=0}^{\infty} \frac{(2l-1)!!}{(2l)!!} \cdot \frac{x^{2l+1}}{2l+1}$, the fourth step follows the fact that $\langle x, y \rangle^{2l} = \langle \Phi^{(2l)}(x), \Phi^{(2l)}(y) \rangle$ where $\Phi^{(2l)}(\cdot)$ is the feature map of the polynomial kernel of degree $2l$ (Definition~\ref{def:gntk_main}).

{\bf Compute kernel matrix $\wt{K}$.}
Finally we compute $\wt{K} \in \R^{n \times n}$. We decompose $\wt{K}$ as follows:
\begin{align*}
    \wt{K} = \wt{K}_1 + \wt{K}_2,
\end{align*}
where we define $\wt{K}_1, \wt{K}_2 \in \R^{n \times n}$ such that for any two graphs ${G_i} = (V_i,E_j)$ and ${G_j} = (V_j, E_j)$,
\begin{align*}
    \wt{K}_1({G_i},{G_j}) = &~ \sum_{u \in V_i, v \in V_j} [\wt{K}^{(1,1)}_1({G_i},{G_j})]_{u,v} \in \R, \\
    \wt{K}_2({G_i},{G_j}) = &~ \sum_{u \in V_i, v \in V_j} [\wt{\Sigma}^{(1,1)}({G_i}, {G_j})]_{u,v} \in \R.
\end{align*}
This equality follows from $\wt{K}({G_i},{G_j}) = \sum_{u \in V_i, v \in V_j} [\wt{K}^{(1,1)}({G_i},{G_j})]_{u,v}$ (see Section~\ref{sec:approx_gntk_formula}) and Eq.~\eqref{eq:pf1_Theta_1_1}. For $\wt{K}_1$, we further have
\begin{align*}
    \wt{K}_1({G_i},{G_j}) = &~ \frac{1}{4} \langle \sum_{u \in V_i} \wt{h}_u, \sum_{v \in V_j} \wt{h}_v \rangle + \frac{1}{2\pi} \sum_{l=1}^{\infty} \frac{(2l-3)!!}{(2l-2)!! (2l-1)} \cdot \langle \sum_{u \in V_i} \Phi^{(2l)}(\wt{h}_u), \sum_{v \in V_j} \Phi^{(2l)}(\wt{h}_v) \rangle \\
    = &~ \frac{1}{4} (\wt{H}^{(1)})^{\top} \cdot \wt{H}^{(1)} + \frac{1}{2\pi} \sum_{l=1}^{\infty} \frac{(2l-3)!!}{(2l-2)!! (2l-1)} \cdot (\wt{H}^{(2l)})^{\top} \cdot \wt{H}^{(2l)},
\end{align*}
where the firs step follows from Eq.~\eqref{eq:pf1_Theta_1_1_1_formula}, the second step follows from the definition of $\wt{H}^{(t)} \in \R^{d \times n}$ (Eq.~\eqref{eq:pf1_wt_H_2l} in Definition~\ref{def:gntk_main}). For $\wt{K}_2$, we have that $\wt{K}_2$ is a kernel matrix, so it is positive semi-definite. 
Let $A \succeq B$ denotes $x^{\top} A x \leq x^{\top} B x$ for all $x$.
Thus we have $\wt{K} \succeq \wt{K}_1$.
\end{proof}

\subsection{Bound on $y^\top \wt{K}^{-1}y$}
\begin{lemma}[Restatement of Lemma~\ref{lem:learnability1_main}]\label{lem:learnability1}
Following the notations of Definition~\ref{def:gntk_main} and under the assumptions of Assumption~\ref{ass:gntk_main}, we have 
\begin{align*}
\| y \|_{ \wt{K}^{-1} } \leq 4 \cdot |\alpha_1| \|\beta_1\|_2 + \sum_{l=1}^{T} 4 \sqrt{\pi} (2l-1) \cdot |\alpha_{2l}| \|\beta_{2l}\|_2.
\end{align*}
\end{lemma}

\begin{proof}
{\bf Decompose $\|y\|_{\wt{K}^{-1}}$.} 
From Part~1 of Assumption~\ref{ass:gntk_main}, we have $\forall i \in [n]$
\begin{align}\label{eq:pf1_y_decompose}
    y_i &= \alpha_1 \sum_{u \in V_i} \langle \ov{h}_u  , \beta_1 \rangle + \sum_{l = 1}^{T}  \alpha_{2l} \sum_{u \in V_i}  \langle \ov{h}_u , \beta_{2l} \rangle ^{2l} \notag \\
    = &~ \alpha_1 \sum_{u \in V_i}  \langle \ov{h}_u , \beta_1 \rangle + \sum_{l = 1}^{T}  \alpha_{2l} \langle \sum_{u \in V_i} \Phi^{(2l)}(\ov{h}_u), \Phi^{(2l)}(\beta_{2l}) \rangle \notag \\
    = &~ y_i^{(1)} + \sum_{l=1}^{T} y_i^{(2l)},
\end{align}
where the second step follows from $\Phi^{(2l)}$ is the feature map of the polynomial kernel of degree $2l$, i.e., $\langle x, y \rangle^{2l} = \langle \Phi^{(2l)}(x), \Phi^{(2l)}(y) \rangle$ for any $x,y \in \R^d$, and the third step follows from defining vectors $y^{(1)}, y^{(2l)} \in \R^n$ for $l \in [T]$ such that $\forall i \in [n]$,
\begin{align}
    y_i^{(1)} := &~ \alpha_1 \langle \sum_{u \in V_i} \ov{h}_u, \beta_1 \rangle \in \R, \label{eq:pf1_y_i_0} \\
    y_i^{(2l)} := &~ \alpha_{2l} \langle \sum_{u \in V_i} \Phi^{(2l)}(\ov{h}_u), \Phi^{(2l)}(\beta_{2l}) \rangle \in \R, ~~ \forall l \in [T], \label{eq:pf1_y_i_2l}
\end{align}
And we have
\begin{align}\label{eq:pf1_y_bound}
    \| y \|_{ \wt{K}^{-1} } \leq\| y \|_{ \wt{K}_1^{-1} }
    \leq \| y^{(1)} \|_{ \wt{K}_1^{-1} } + \sum_{l=1}^{T} \| y^{(2l)} \|_{\wt{K}_1^{-1}},
\end{align}
which follows from Lemma~\ref{lem:close_form_approx_gntk} that $\wt{K} \succeq \wt{K}_1$ and thus $\wt{K}^{-1} \preceq \wt{K}_1^{-1}$, the second step follows from Eq.~\eqref{eq:pf1_y_decompose} and triangle inequality.

{\bf Upper bound $\|y^{(1)}\|_{\wt{K}_1^{-1}}$.}
Recall the definitions of the two matrices $\ov{H}^{(1)}, \wt{H}^{(1)} \in \R^{d \times n}$ in Eq.~\eqref{eq:pf1_ov_H_2l} and \eqref{eq:pf1_wt_H_2l} of Definition~\ref{def:gntk_main}:
\begin{align*}
    \ov{H}^{(1)} := \Big[ \sum_{u \in V_1} \ov{h}_u, \sum_{u \in V_2} \ov{h}_u, \cdots, \sum_{u \in V_n} \ov{h}_u \Big], ~~~
    \wt{H}^{(1)} := \Big[ \sum_{u \in V_1} \wt{h}_u, \sum_{u \in V_2} \wt{h}_u, \cdots, \sum_{u \in V_n} \wt{h}_u \Big].
\end{align*}
Note that from Eq.~\eqref{eq:pf1_y_i_0} we have $y^{(1)} = \alpha_1 \cdot (\ov{H}^{(1)})^{\top} \beta_1 \in \R^{n}$. Also, from Lemma~\ref{lem:close_form_approx_gntk} we have $\wt{K}_1 \succeq \frac{1}{4} (\wt{H}^{(1)})^{\top} \wt{H}^{(1)} \in \R^{n \times n}$. Using these two equations, we have
\begin{align}\label{eq:pf1_y_1_Theta_1_bound}
    \|y^{(1)}\|_{\wt{K}_1^{-1}}^2 = &~ (y^{(1)})^{\top} \wt{K}^{-1}_{1} y^{(1)} \notag \\
    \leq &~ 4 \alpha_1^2 \cdot \beta_1^{\top} \ov{H}^{(1)} \cdot ((\wt{H}^{(1)})^{\top} \wt{H}^{(1)})^{-1} \cdot (\ov{H}^{(1)})^{\top} \beta_1.
\end{align}

Next we want to prove that $(1-\frac{1}{2}) (\ov{H}^{(1)})^{\top} \ov{H}^{(1)} \preceq (\wt{H}^{(1)})^{\top} \wt{H}^{(1)} \preceq (1+\frac{1}{2}) (\ov{H}^{(1)})^{\top} \ov{H}^{(1)}$. For any $i, j \in [n]$, we have
\begin{align*}
    [(\ov{H}^{(1)})^{\top} \ov{H}^{(1)}]_{i,j} = &~ (\sum_{u \in V_i} \ov{h}_u^{\top}) \cdot (\sum_{v \in V_j} \ov{h}_v)
    = \mathbf{1}_{[N_i]}^{\top} C_{G_i}^{\top} A_{G_i}^{\top} H_{G_i}^{\top} \cdot H_{G_j} A_{G_j} C_{G_j} \mathbf{1}_{[N_j]}, \\
    [(\wt{H}^{(1)})^{\top} \wt{H}^{(1)}]_{i,j} = &~ (\sum_{u \in V_i} \wt{h}_u^{\top}) \cdot (\sum_{v \in V_j} \wt{h}_v)
    = \mathbf{1}_{[N_i]}^{\top} C_{G_i}^{\top} A_{G_i}^{\top} (S_{G_i}^{\top} S_{G_i}) H_{G_i}^{\top} \cdot H_{G_j} (S_{G_j}^{\top} S_{G_j}) A_{G_j} C_{G_j} \mathbf{1}_{[N_j]},
\end{align*}
where $\mathbf{1}_{[N_i]} \in \R^{N_i}$ is the all one vector, and the second steps of the two equations follow from $\ov{h}_u = [H_{G_i} A_{G_i} C_{G_i}]_{*,u}$ and $\wt{h}_u = [H_{G_i} S_{G_i}^{\top} S_{G_i} A_{G_i} C_{G_i}]_{*,u}$ (see Definition~\ref{def:gntk_main}).

For any $i,j \in [n]$, using Lemma~\ref{lem:CE_two_sketch} we have that with probability $1 - 1/N^4$,
\begin{align}\label{eq:pf1_wt_H_minus_ov_H_ij}
    &~ |[(\wt{H}^{(1)})^{\top} \wt{H}^{(1)}]_{i,j} - [(\ov{H}^{(1)})^{\top} \ov{H}^{(1)}]_{i,j}| \notag \\
    \leq &~ \frac{O(\log^{1.5} N)}{\sqrt{b_i}} \cdot \|A_{G_i} C_{G_i} \mathbf{1}_{[N_i]}\|_2 \cdot \|H_{G_i}^{\top} H_{G_j} A_{G_j} C_{G_j} \mathbf{1}_{[N_j]}\|_2 \notag \\
    &~ + \frac{O(\log^{1.5} N)}{\sqrt{b_j}} \cdot \|A_{G_j} C_{G_j} \mathbf{1}_{[N_j]}\|_2 \cdot \|H_{G_j}^{\top} H_{G_i} A_{G_i} C_{G_i} \mathbf{1}_{[N_i]}\|_2 \notag \\
    &~ + \frac{O(\log^{3} N)}{\sqrt{b_i b_j}} \cdot \|A_{G_j} C_{G_j} \mathbf{1}_{[N_j]}\|_2 \|A_{G_i} C_{G_i} \mathbf{1}_{[N_i]}\|_2 \cdot \|H_{G_j}^{\top} H_{G_i}\|_F \notag \\
    \leq &~ \frac{1}{10 \gamma T} \cdot [(\ov{H}^{(1)})^{\top} \ov{H}^{(1)}]_{i,j},
\end{align}
where the last step follows Part~3 of Assumption~\ref{ass:gntk_main} that
\begin{align*}
    &~ \|A_{G_j} C_{G_j} \mathbf{1}_{[N_j]}\|_2 \|A_{G_i} C_{G_i} \mathbf{1}_{[N_i]}\|_2 \cdot \|H_{G_j}^{\top} H_{G_i}\|_F \\
    \leq &~ O(1) \cdot \frac{\min\{\sqrt{b_i}, \sqrt{b_j}\}}{\gamma T \log^3 N} \cdot \mathbf{1}_{[N_i]}^{\top} C_{G_i}^{\top} A_{G_i}^{\top} H_{G_i}^{\top} H_{G_j} A_{G_j} C_{G_j} \mathbf{1}_{[N_j]}.
\end{align*}

Then we have that
\begin{align*}
    \|(\wt{H}^{(1)})^{\top} \wt{H}^{(1)} - (\ov{H}^{(1)})^{\top} \ov{H}^{(1)}\|_2^2 \leq &~ \|(\wt{H}^{(1)})^{\top} \wt{H}^{(1)} - (\ov{H}^{(1)})^{\top} \ov{H}^{(1)}\|_F^2 \\
    = &~ \sum_{i=1}^n \sum_{j=1}^n |[(\wt{H}^{(1)})^{\top} \wt{H}^{(1)}]_{i,j} - [(\ov{H}^{(1)})^{\top} \ov{H}^{(1)}]_{i,j}|^2 \\
    \leq &~ \frac{1}{100 \gamma^2 T^2} \sum_{i=1}^n \sum_{j=1}^n [(\ov{H}^{(1)})^{\top} \ov{H}^{(1)}]_{i,j}^2 \\
    = &~ \frac{1}{100 \gamma^2 T^2} \|(\ov{H}^{(1)})^{\top} \ov{H}^{(1)}\|_F^2 \\
    \leq &~ \frac{1}{100 T^2} \|(\ov{H}^{(1)})^{\top} \ov{H}^{(1)}\|_2^2,
\end{align*}
where the third step uses Eq.~\eqref{eq:pf1_wt_H_minus_ov_H_ij}, the fifth step follows Part~2 of Assumption~\ref{ass:gntk_main} that $\|(\ov{H}^{(1)})^{\top} \ov{H}^{(1)}\|_F \leq \gamma \cdot \|(\ov{H}^{(1)})^{\top} \ov{H}^{(1)}\|_2$.

Thus we have proven that
\begin{align*}
    (1-\frac{1}{10}) (\ov{H}^{(1)})^{\top} \ov{H}^{(1)} \preceq (\wt{H}^{(1)})^{\top} \wt{H}^{(1)} \preceq (1+\frac{1}{10}) (\ov{H}^{(1)})^{\top} \ov{H}^{(1)}.
\end{align*}

Using this fact, we can bound $\|y^{(1)}\|_{\wt{K}_1^{-1}}$ as follows:
\begin{align}\label{eq:pf1_y_1_Theta_1_final}
    \|y^{(1)}\|_{\wt{K}_1^{-1}}
    \leq &~ (4 \alpha_1^2 \cdot \beta_1^{\top} \ov{H}^{(1)} \cdot ((\wt{H}^{(1)})^{\top} \wt{H}^{(1)})^{-1} \cdot (\ov{H}^{(1)})^{\top} \beta_1)^{1/2} \notag \\
    \leq &~ (8 \alpha_1^2 \cdot \beta_1^{\top} \ov{H}^{(1)} \cdot ((\ov{H}^{(1)})^{\top} \ov{H}^{(1)})^{-1} \cdot (\ov{H}^{(1)})^{\top} \beta_1)^{1/2} \notag \\
    \leq &~ 4 \cdot \alpha_1 \|\beta_1\|_2.
\end{align}
where the first step uses Eq.~\eqref{eq:pf1_y_1_Theta_1_bound}.

{\bf Upper bound $\|y^{(2l)}\|_{\wt{K}_1^{-1}}$.} Consider some $l \in [T]$.
Recall the definitions of the two matrices $\ov{H}^{(2l)}, \wt{H}^{(2l)} \in \R^{d \times n}$ in Eq.~\eqref{eq:pf1_ov_H_2l} and \eqref{eq:pf1_wt_H_2l} of Definition~\ref{def:gntk_main}:
\begin{align*}
    \ov{H}^{(2l)} := &~ \Big[ \sum_{u \in V_1} \Phi^{(2l)}(\ov{h}_u), \sum_{u \in V_2} \Phi^{(2l)}(\ov{h}_u), \cdots, \sum_{u \in V_n} \Phi^{(2l)}(\ov{h}_u) \Big], \\
    \wt{H}^{(2l)} := &~ \Big[ \sum_{u \in V_1} \Phi^{(2l)}(\wt{h}_u), \sum_{u \in V_2} \Phi^{(2l)}(\wt{h}_u), \cdots, \sum_{u \in V_n} \Phi^{(2l)}(\wt{h}_u) \Big].
\end{align*}
Note that from Eq.~\eqref{eq:pf1_y_i_2l} we have $y^{(2l)} = \alpha_{2l} \cdot (\ov{H}^{(2l)})^{\top} \Phi^{(2l)}(\beta_{2l}) \in \R^{n}$. Also, from Lemma~\ref{lem:close_form_approx_gntk} we have $\wt{K}_1 \succeq \frac{(2l-3)!!}{2 \pi (2l-2)!! (2l-1)} (\wt{H}^{(2l)})^{\top} \wt{H}^{(2l)} \in \R^{n \times n}$.
Using these two equations, we have
\begin{align}\label{eq:pf1_y_2l_Theta_1_bound}
    \|y^{(2l)}\|_{\wt{K}_1^{-1}}^2 = &~ (y^{(2l)} )^{\top} \wt{K}^{-1}_{1} y^{(2l)} \notag \\
    \leq &~ \frac{2 \pi (2l-2)!! (2l-1)}{(2l-3)!!} \alpha_{2l}^2 \cdot \beta_{2l}^{\top} \ov{H}^{(2l)} \cdot ((\wt{H}^{(2l)})^{\top} \wt{H}^{(2l)})^{-1} \cdot (\ov{H}^{(2l)})^{\top} \beta_{2l}.
\end{align}

Next we want to prove that $(1-\frac{1}{2}) (\ov{H}^{(2l)})^{\top} \ov{H}^{(2l)} \preceq (\wt{H}^{(2l)})^{\top} \wt{H}^{(2l)} \preceq (1+\frac{1}{2}) (\ov{H}^{(2l)})^{\top} \ov{H}^{(2l)}$. For any $i, j \in [n]$, we have
\begin{align}
    [(\ov{H}^{(2l)})^{\top} \ov{H}^{(2l)}]_{i,j} = &~ (\sum_{u \in V_i} \Phi^{(2l)}(\ov{h}_u)^{\top}) \cdot (\sum_{v \in V_j} \Phi^{(2l)}(\ov{h}_v))
    = \sum_{u \in V_i} \sum_{v \in V_j} (\ov{h}_u^{\top} \ov{h}_v)^{2l} \label{eq:pf1_ov_H_2l_ij} \\
    [(\wt{H}^{(2l)})^{\top} \wt{H}^{(2l)}]_{i,j} = &~ (\sum_{u \in V_i} \Phi^{(2l)}(\wt{h}_u)^{\top}) \cdot (\sum_{v \in V_j} \Phi^{(2l)}(\wt{h}_v))
    = \sum_{u \in V_i} \sum_{v \in V_j} (\wt{h}_u^{\top} \wt{h}_v)^{2l} \label{eq:pf1_wt_H_2l_ij}
\end{align}
where the second steps of both equations follow from $\Phi^{(2l)}(x)^{\top} \Phi^{(2l)}(y) = (x^{\top} y)^{2l}$.

Note that in Eq.~\eqref{eq:pf1_wt_H_minus_ov_H_ij} we have proven that
\begin{align*}
    |\sum_{u \in V_i} \sum_{v \in V_j} (\ov{h}_u^{\top} \ov{h}_v - \wt{h}_u^{\top} \wt{h}_v)| \leq \frac{1}{10 \gamma T} \sum_{u \in V_i} \sum_{v \in V_j} \ov{h}_u^{\top} \ov{h}_v.
\end{align*}
Thus we have
\begin{align}\label{eq:pf1_wt_H_minus_ov_H_ij_2l_helpful}
    \big|\sum_{u \in V_i} \sum_{v \in V_j} \big((\ov{h}_u^{\top} \ov{h}_v)^{2l} - (\wt{h}_u^{\top} \wt{h}_v)^{2l} \big) \big| = &~ \big|\sum_{u \in V_i} \sum_{v \in V_j} \sum_{i=0}^{2l-1} (\ov{h}_u^{\top} \ov{h}_v - \wt{h}_u^{\top} \wt{h}_v) (\ov{h}_u^{\top} \ov{h}_v)^i (\wt{h}_u^{\top} \wt{h}_v)^{2l-1-i} \big| \notag \\
    \leq &~ 2l \cdot |\sum_{u \in V_i} \sum_{v \in V_j} (\ov{h}_u^{\top} \ov{h}_v - \wt{h}_u^{\top} \wt{h}_v)| \cdot (\max_{u,v}\{|\ov{h}_u^{\top} \ov{h}_v|, |\wt{h}_u^{\top} \wt{h}_v|\})^{2l-1} \notag \\
    \leq &~ 2l \cdot \frac{1}{10 \gamma T} \cdot (1 + \frac{1}{10 \gamma T})^{2l} \cdot (\sum_{u \in V_i} \sum_{v \in V_j} \ov{h}_u^{\top} \ov{h}_v)^{2l} \notag \\
    \leq &~ \frac{1}{2 \gamma} (\sum_{u \in V_i} \sum_{v \in V_j} \ov{h}_u^{\top} \ov{h}_v)^{2l},
\end{align}
where the first step follows from $x^{2l} - y^{2l} = \sum_{i=0}^{2l-1} (x-y) x^i y^{2l-1-i}$, the third step follows from $\max_{u,v}\{|\ov{h}_u^{\top} \ov{h}_v|\} \leq \sum_{u \in V_i} \sum_{v \in V_j} \ov{h}_u^{\top} \ov{h}_v$ since $h_u$ and $h_v$ are non-negative, and the previous inequality from Eq.~\eqref{eq:pf1_wt_H_minus_ov_H_ij}, the fourth step follows from $l \leq T$ and $(1+\frac{1}{10 \gamma T})^{2l} \leq (1+\frac{1}{10 l})^{2l} \leq \sqrt{e}$.

Thus we have proven that for any $i, j \in [n]$, we have
\begin{align}\label{eq:pf1_wt_H_minus_ov_H_ij_2l}
    \big|[(\wt{H}^{(2l)})^{\top} \wt{H}^{(2l)}]_{i,j} - [(\ov{H}^{(2l)})^{\top} \ov{H}^{(2l)}]_{i,j}\big|
    = &~ |\sum_{u \in V_i} \sum_{v \in V_j} (\ov{h}_u^{\top} \ov{h}_v)^{2l} - \sum_{u \in V_i} \sum_{v \in V_j} (\wt{h}_u^{\top} \wt{h}_v)^{2l}| \notag \\
    \leq &~ \frac{1}{2 \gamma} [(\ov{H}^{(2l)})^{\top} \ov{H}^{(2l)}]_{i,j}.
\end{align}
where the first step uses Eq.~\eqref{eq:pf1_ov_H_2l_ij} and \eqref{eq:pf1_wt_H_2l_ij}, the second step uses Eq.~\eqref{eq:pf1_wt_H_minus_ov_H_ij_2l_helpful}.

Now similar to how we bound $\|y^{(1)}\|_{\wt{K}_1^{-1}}$, we have that
\begin{align*}
    \|(\wt{H}^{(2l)})^{\top} \wt{H}^{(2l)} - (\ov{H}^{(2l)})^{\top} \ov{H}^{(2l)}\|_2^2 \leq &~ \|(\wt{H}^{(2l)})^{\top} \wt{H}^{(2l)} - (\ov{H}^{(2l)})^{\top} \ov{H}^{(2l)}\|_F^2 \\
    = &~ \sum_{i=1}^n \sum_{j=1}^n |[(\wt{H}^{(2l)})^{\top} \wt{H}^{(2l)}]_{i,j} - [(\ov{H}^{(2l)})^{\top} \ov{H}^{(2l)}]_{i,j}|^2 \\
    \leq &~ \frac{1}{4 \gamma^2} \sum_{i=1}^n \sum_{j=1}^n [(\ov{H}^{(2l)})^{\top} \ov{H}^{(2l)}]_{i,j}^2 \\
    = &~ \frac{1}{4 \gamma^2} \|(\ov{H}^{(2l)})^{\top} \ov{H}^{(2l)}\|_F^2 \\
    \leq &~ \frac{1}{4} \|(\ov{H}^{(2l)})^{\top} \ov{H}^{(2l)}\|_2^2,
\end{align*}
where the third step uses Eq.~\eqref{eq:pf1_wt_H_minus_ov_H_ij_2l}, the fifth step follows Part~2 of Assumption~\ref{ass:gntk_main} that $\|(\ov{H}^{(2l)})^{\top} \ov{H}^{(2l)}\|_F \leq \gamma \cdot \|(\ov{H}^{(2l)})^{\top} \ov{H}^{(2l)}\|_2$.

Thus we have proven that
\begin{align*}
    (1-\frac{1}{2}) (\ov{H}^{(2l)})^{\top} \ov{H}^{(2l)} \preceq (\wt{H}^{(2l)})^{\top} \wt{H}^{(2l)} \preceq (1+\frac{1}{2}) (\ov{H}^{(2l)})^{\top} \ov{H}^{(2l)}.
\end{align*}

Using this fact, we can bound $\|y^{(2l)}\|_{\wt{K}_1^{-1}}$ as follows:
\begin{align}\label{eq:pf1_y_2l_Theta_1_final}
    \|y^{(2l)}\|_{\wt{K}_1^{-1}}
    \leq &~ \Big( \frac{2 \pi (2l-2)!! (2l-1)}{(2l-3)!!} \alpha_{2l}^2 \cdot \beta_{2l}^{\top} \ov{H}^{(2l)} \cdot ((\wt{H}^{(2l)})^{\top} \wt{H}^{(2l)})^{-1} \cdot (\ov{H}^{(2l)})^{\top} \beta_{2l} \Big)^{1/2} \notag \\
    \leq &~ \Big( \frac{8 \pi (2l-2)!! (2l-1)}{(2l-3)!!} \alpha_{2l}^2 \cdot \beta_{2l}^{\top} \ov{H}^{(2l)} \cdot ((\wt{H}^{(2l)})^{\top} \wt{H}^{(2l)})^{-1} \cdot (\ov{H}^{(2l)})^{\top} \beta_{2l} \Big)^{1/2} \notag \\
    \leq &~ 4 \sqrt{\pi} (2l-1) \cdot \alpha_{2l} \|\beta_{2l}\|_2,
\end{align}
where the first step uses Eq.~\eqref{eq:pf1_y_2l_Theta_1_bound}.

{\bf Upper bound $\|y\|_{\wt{K}^{-1}}$.}
Plugging Eq.~\eqref{eq:pf1_y_1_Theta_1_final} and \eqref{eq:pf1_y_2l_Theta_1_final} into Eq.~\eqref{eq:pf1_y_bound}, we have
\begin{align*}
    \| y \|_{ \wt{K}^{-1} } \leq 4 \cdot \alpha_1 \|\beta_1\|_2 + \sum_{l=1}^{T} 4 \sqrt{\pi} (2l-1) \cdot \alpha_{2l} \|\beta_{2l}\|_2.
\end{align*}
\end{proof}

\subsection{Bound on trace of $\wt{K}$}
\begin{lemma}[Restatement of Lemma~\ref{lem:trace_main}]\label{lem:trace}
Following the notations of Definition~\ref{def:gntk_main} and under the assumptions of Assumption~\ref{ass:gntk_main}, we have
\begin{align*}
    \tr[\wt{K}] \leq 2 n N^2.
\end{align*}
\end{lemma}
\begin{proof}
From Lemma~\ref{lem:close_form_approx_gntk} we can decompose $\wt{K} \in \R^{n \times n}$ into
\begin{align*}
    \wt{K} = \wt{K}_1 + \wt{K}_2 \succeq \wt{K}_1.
\end{align*}
And for each $i \in [n]$, Lemma~\ref{lem:close_form_approx_gntk} gives the following bound on the diagonal entries of $\wt{K}_1$:
\begin{align*}
    \wt{K}_1({G_i},{G_i}) = &~ \sum_{u \in V_i} \sum_{v \in V_i} \langle \wt{h}_u, \wt{h}_v \rangle \cdot \frac{1}{2\pi} \big(\pi - \arccos (\langle \wt{h}_u, \wt{h}_v \rangle) \big) \\
    \leq &~ \sum_{u \in V_i} \sum_{v \in V_i} \frac{1}{2} = \frac{N_i^2}{2},
\end{align*}
where the second step follows from $\wt{h}_u$ and $\wt{h}_v$ are unit vectors and that $\arccos(\cdot) \in [0,\pi]$.

Lemma~\ref{lem:close_form_approx_gntk} also gives the following bound on the diagonal entries of $\wt{K}_2$:
\begin{align*}
    \wt{K}_2({G_i},{G_i}) = &~ \sum_{u \in V_i} \sum_{v \in V_i} \langle \wt{h}_u, \wt{h}_v \rangle \cdot \frac{1}{2\pi} \Big(\pi - \arccos (\langle \wt{h}_u, \wt{h}_v \rangle) + \sqrt{1 - \langle \wt{h}_u, \wt{h}_v \rangle^2} \Big) \\
    \leq &~ \sum_{u \in V_i} \sum_{v \in V_i} \frac{1}{2\pi} (\pi + 1) \leq N_i^2.
\end{align*}
where the second step follows from $\wt{h}_u$ and $\wt{h}_v$ are unit vectors and that $\arccos(\cdot) \in [0,\pi]$.

Thus we have
\begin{align*}
    \tr[\wt{K}] = \tr[\wt{K}_1] + \tr[\wt{K}_2] \leq 2 \sum_{i=1}^n N_i^2 \leq 2 n N^2. 
\end{align*}
\end{proof}

%% file: gntk_time.tex
\section{Running time}\label{sec:time}
\begin{theorem}[Main running time theorem]\label{thm:time_main_formal}
Consider a GNN with $L$ levels of BLOCK operations, and $R$ hidden layers in each level. We compute the kernel matrix using $n$ graphs $G_1=(V_1,E_1), \cdots, G_n=(V_n,E_n)$ with $|V_i| = N_i$. Let $b_i \leq N_i$ be the sketch size of $G_i$. Let $d \in \mathbb{N}_+$ be the dimension of the feature vectors.

The total running time to compute the approximate GNTK is
\begin{align*}
    \sum_{i=1}^n \sum_{j=1}^n \Tmat(N_i, d, N_j) + O(L) \cdot \sum_{i=1}^n \sum_{j=1}^n \Tmat(N_i, N_j, b_i) + O(LR) \cdot (\sum_{i=1}^n N_i)^2.
\end{align*}

When assuming $N_i \leq N$ and $b_i \leq b$ for all $i \in [n]$, the total running time is
\begin{align*}
    O(n^2) \cdot (\Tmat(N, N, d) + L \cdot \Tmat(N, N, b) + LR \cdot N^2).
\end{align*}
\end{theorem}
\begin{proof}

{\bf Preprocessing time.} When preprocessing, we compute $A_{G_i} S_{G_i}^{\top}$ for each $i \in [n]$ in $\Tmat(N_i, N_i, b_i)$ time. So in total we need $\sum_{i=1}^n \Tmat(N_i, N_i, b_i)$ time.

We also compute the initial matrices $\wt{\Sigma}^{(0,R)}(G_i,G_j), \wt{K}^{(0,R)}(G_i,G_j) \in \R^{N_i \times N_j}$ for each $i,j \in [n]$ with $[\wt{\Sigma}^{(0,R)}(G_i,G_j)]_{u,v} = [K^{(0,R)}(G_i,G_j)]_{u,v} = \langle h_u, h_v \rangle$, $\forall u \in V_i, v \in V_j$. Computing each $\wt{\Sigma}^{(0,R)}(G_i,G_j)$ corresponds to multiplying the concatenation of the feature vectors of $G_i$ with that of $G_j$, which is multiplying a $N_i \times d$ matrix with a $d \times N_j$ matrix, and this takes $\Tmat(N_i, d, N_j)$ time. So computing all initial matrices takes $\sum_{i=1}^n \sum_{j=1}^n \Tmat(N_i, d, N_j)$ time.

Thus the total preprocessing time is
\begin{align*}
    \sum_{i=1}^n \Tmat(N_i, N_i, b_i) + \sum_{i=1}^n \sum_{j=1}^n \Tmat(N_i, d, N_j).
\end{align*}

{\bf {\block} operation: aggregation time.}
In the $l$-th level of {\block} operation, $\forall i,j \in [n]$ we compute $\wt{\Sigma}^{(l,0)}(G_i,G_j), \wt{K}^{(l,0)}(G_i,G_j) \in \R^{N_i \times N_j}$ by computing 
\begin{align*}
\wt{\Sigma}^{(\ell,0)}(G_i,G_j) := &~ C_{G_i} (A_{G_i} S_{G_i}^{\top}) \cdot S_{G_i} \cdot \wt{\Sigma}^{(\ell-1, R)}(G_i,G_j) \cdot S_{G_j}^{\top} \cdot (S_{G_j} A_{G_j}) C_{G_j}, \\
\wt{K}^{(\ell,0)}(G,H) := &~ C_{G_i} (A_{G_i} S_{G_i}^{\top}) \cdot S_{G_i} \cdot \wt{K}^{(\ell-1, R)}(G_i,G_j) \cdot S_{G_j}^{\top} \cdot (S_{G_j} A_{G_j}) C_{G_j}.
\end{align*}
This takes $O(\Tmat(N_i, N_j, b_i) + \Tmat(N_i, N_j, b_j))$ time.

Thus the total time of all aggregation operations of $L$ levels is
\begin{align*}
    O(L) \cdot \sum_{i=1}^n \sum_{j=1}^n \Tmat(N_i, N_j, b_i).
\end{align*}

{\bf {\block} operation: hidden layer time.}
In the $l$-th level of {\block} operation, in the $r$-th hidden layer, for each $i,j \in [n]$ we compute $\wt{\Sigma}^{(l,r)}(G_i,G_j), \wt{K}^{(l,r)}(G_i,G_j) \in \R^{N_i \times N_j}$ by computing each entry $[\wt{\Sigma}^{(l,r)}(G_i,G_j)]_{u,v}, [\wt{K}^{(l,r)}(G_i,G_j)]_{u,v} \in \R$ for $u \in V_i$, $v \in V_j$. Computing each entry takes $O(1)$ time, which follows trivially from their definitions (see Section~\ref{sec:our_gntk_formulation} and \ref{sec:approx_gntk_formula}).
Thus the total time of all $R$ hidden layers operations of $L$ levels is
\begin{align*}
    O(LR) \cdot (\sum_{i=1}^n N_i)^2.
\end{align*}

{\bf {\readout} operation time.} Finally we compute kernel matrix $K \in \R^{n \times n}$ such that for $i,j \in [n]$, $K(G_i,G_j) \in \R$ is computed as
\begin{align*}
	K(G_i,G_j) =
	\sum_{u \in V_i, v \in V_j}  [K^{(L, R)}(G_i, G_j) ]_{u, v}.
\end{align*}
Thus the total time of {\readout} operation is 
\begin{align*}
    (\sum_{i=1}^n N_i)^2.
\end{align*}

{\bf Total time.} Thus the total running time to compute the approximate GNTK is
\begin{align*}
    \sum_{i=1}^n \sum_{j=1}^n \Tmat(N_i, d, N_j) + O(L) \cdot \sum_{i=1}^n \sum_{j=1}^n \Tmat(N_i, N_j, b_i) + O(LR) \cdot (\sum_{i=1}^n N_i)^2.
\end{align*}

When assuming $N_i \leq N$ and $b_i \leq b$ for all $i \in [n]$, the total running time is
\begin{align*}
    O(n^2) \cdot (\Tmat(N, N, d) + L \cdot \Tmat(N, N, b) + LR \cdot N^2).
\end{align*}
\end{proof}

For comparison we state the running time of computing GNTK of \cite{dhs+19}.
\begin{theorem}[Running time of \cite{dhs+19}]\label{thm:time_dhs}
Consider a GNN with $L$ levels of BLOCK operations, and $R$ hidden layers in each level. We compute the kernel matrix using $n$ graphs $G_1=(V_1,E_1), \cdots, G_n=(V_n,E_n)$ with $|V_i| = N_i$. Let $d \in \mathbb{N}_+$ be the dimension of the feature vectors.

The total running time of \cite{dhs+19} to compute the GNTK is
\begin{align*}
    \sum_{i=1}^n \sum_{j=1}^n \Tmat(N_i, d, N_j) + O(L) \cdot (\sum_{i=1}^n N_i^2)^2 + O(LR) \cdot (\sum_{i=1}^n N_i)^2.
\end{align*}

When assuming $N_i \leq N$ and $b_i \leq b$ for all $i \in [n]$, the total running time is
\begin{align*}
    O(n^2) \cdot (\Tmat(N, N, d) + L \cdot N^4 + LR \cdot N^2).
\end{align*}
\end{theorem}
We include a proof here for completeness.
\begin{proof}
Comparing with Theorem~\ref{thm:time_main_formal}, the only different part of the running time is the aggregation time of {\block} operation. For the other three parts, see the proof of Theorem~\ref{thm:time_main_formal}.

{\bf {\block} operation: aggregation time.}
In the $l$-th level of {\block} operation, $\forall i,j \in [n]$ we compute ${\Sigma}^{(l,0)}(G_i,G_j), {K}^{(l,0)}(G_i,G_j) \in \R^{N_i \times N_j}$ by computing 
\begin{align*}
\vect({\Sigma}^{(\ell,0)}(G_i,G_j)) := &~ ((C_{G_i} A_{G_i}) \otimes (C_{G_j} A_{G_j})) \cdot \vect({\Sigma}^{(\ell-1, R)}(G_i,G_j)) \in \R^{N_i N_j} , \\
\vect({K}^{(\ell,0)}(G_i,G_j)) := &~ ((C_{G_i} A_{G_i}) \otimes (C_{G_j} A_{G_j})) \cdot \vect({K}^{(\ell-1, R)}(G_i,G_j)) \in \R^{N_i N_j}. \\
\end{align*}
Note that the sizes are $((C_{G_i} A_{G_i}) \otimes (C_{G_j} A_{G_j})) \in \R^{N_i N_j \times N_i N_j}$, $\vect({\Sigma}^{(\ell-1, R)}(G_i,G_j)) \in \R^{N_i N_j}$.
So this takes $O(N_i^2 N_j^2)$ time, even to simply compute $((C_{G_i} A_{G_i}) \otimes (C_{G_j} A_{G_j}))$.

Thus the total time of all aggregation operations of $L$ levels is
\begin{align*}
    O(L) \cdot (\sum_{i=1}^n N_i^2)^2.
\end{align*}

{\bf Total time.} Thus the total running time in \cite{dhs+19} to compute the exact GNTK is
\begin{align*}
    \sum_{i=1}^n \sum_{j=1}^n \Tmat(N_i, d, N_j) + O(L) \cdot (\sum_{i=1}^n N_i^2)^2 + O(LR) \cdot (\sum_{i=1}^n N_i)^2.
\end{align*}

When assuming $N_i \leq N$ and $b_i \leq b$ for all $i \in [n]$, the total running time is
\begin{align*}
    O(n^2) \cdot (\Tmat(N, N, d) + L \cdot N^4 + LR \cdot N^2).
\end{align*}
\end{proof}

%% file: app_sketching.tex
\section{Missing proofs for Kronecker product and Sketching}\label{sec:kron_sketching_proof}
\subsection{Proofs of Kronecker product equivalence}

\begin{fact}[Equivalence between two matrix products and Kronecker product then matrix-vector multiplication]\label{fac:app_equivalence_matrix_product_kronecker}
Given matrices $A \in \R^{n_1 \times d_1}$, $B \in \R^{n_2 \times d_2}$, and $H \in \R^{d_1 \times d_2}$, we have $\vect(A H B^\top) = (A \otimes B) \cdot \vect(H)$.
\end{fact}
\begin{proof}
First note that $A H B^{\top} \in \R^{n_1 \times n_2}$, $A \otimes B \in \R^{n_1 n_2 \times d_1 d_2}$, and $(A \otimes B) \cdot \vect(H) \in \R^{n_1 n_2}$.

For any $i_1 \in [n_1]$, $i_2 \in [n_2]$, define $i := i_1 + (i_2 - 1) \cdot n_1$, we have
\begin{align*}
    \vect(A H B^{\top})_i = &~ (A H B^{\top})_{i_1, i_2} \\
    = &~ \sum_{j_1 \in [d_1]} \sum_{j_2 \in [d_2]} A_{i_1, j_1} \cdot H_{j_1, j_2} \cdot B_{i_2, j_2},
\end{align*}
and we also have,
\begin{align*}
    \big( (A \otimes B) \cdot \vect(H) \big)_{i} = &~ \sum_{\substack{j := j_1 + (j_2 - 1) \cdot d_1, \\ j_1 \in [d_1], j_2 \in [d_2] }} (A \otimes B)_{i, j} \cdot \vect(H)_j \\
    = &~ \sum_{j_1 \in [d_1], j_2 \in [d_2]} A_{i_1, j_1} B_{i_2, j_2} \cdot H_{j_1, j_2}.
\end{align*}

Thus we have $\vect(A H B^\top) = (A \otimes B) \cdot  \vect(H)$.
\end{proof}

\subsection{Proof of sketching bound}\label{sec:app_sketching}

We will use the following inequality.
\begin{fact}[Khintchine’s inequality]\label{fac:khintchine}
Let $\sigma_1,\sigma_2,\cdots,\sigma_n$ be i.i.d. sign random variables, and let $z_1,z_2,\cdots,z_n \in \R$. Then there exist constants $C,C' > 0$ such that $\forall t \in \R_+$,
\begin{align*}
    \Pr\Big[ \Big| \sum_{i=1}^n \sigma_i z_i \Big| \geq C t \|z\|_2 \Big] \leq e^{-C' t^2}.
\end{align*}
\end{fact}

\begin{lemma}[Restatement of Lemma~\ref{lem:CE_two_sketch}]
Let $A \in \R^{n \times n}$ be a matrix. Let $R \in \R^{b_1 \times n}$ and $S \in \R^{b_2 \times n}$ be two independent AMS matrices. Let $g, h \in \R^n$ be two vectors. Then with probability at least $1-\poly(1/n)$, we have
\begin{align*}
    &~ g^{\top} (R^{\top} R) A (S^{\top} S) h - g^{\top} A h \\
    \leq &~ O(\frac{\log^{1.5} n}{\sqrt{b_1}}) \|g\|_2 \|A h\|_2 + O(\frac{\log^{1.5} n}{\sqrt{b_2}}) \|g^{\top} A\|_2 \|h\|_2 + O(\frac{\log^{3} n}{\sqrt{b_1 b_2}}) \cdot \|g\|_2 \|h\|_2 \|A\|_F.
\end{align*}
\end{lemma}
\begin{proof}
For $i\in [n]$, we use $R_i \in \R^{b_1}$ and $S_i \in \R^{b_2}$ to denote the $i$-th column of $R$ and $S$. 

Each column $R_i$ of the AMS matrix $R$ has the same distribution as $\sigma_i R_i$, where $\sigma_i$ is a random sign. The AMS matrix $R$ has the following properties:
\begin{align}
    1. &~ \langle R_i, R_i\rangle = 1, \forall i \in [n]. \label{eq:two_sketch_same_Ri} \\
    2. &~ \Pr \Big[\langle R_i, R_j\rangle \leq \frac{\sqrt{\log(n/\delta)}}{\sqrt{b_1}}, \forall i\neq j \in [n] \Big] \geq 1-\delta. \label{eq:two_sketch_diff_Ri}
\end{align}
Similarly each column $S_i$ of AMS matrix $S$ has the same distribution as $\sigma'_{i} S_i$, where $\sigma'_i$ is a random sign. For more details see \cite{ams99}.

We have
\begin{align}\label{eq:two_sketch_close_form_whp}
    g^{\top} (R^{\top} R) A (S^{\top} S) h = &~ \sum_{i,j,i',j'} g_{i} h_{j'} \sigma_{i} \sigma_{j} \sigma'_{i'} \sigma'_{j'} \langle R_i, R_j \rangle A_{j,i'} \langle S_{i'}, S_{j'} \rangle.
\end{align}
Thus we can split the summation of Eq.~\eqref{eq:two_sketch_close_form_whp} into three parts: 1. Two pairs of indexes are the same: $i = j$ and $i' = j'$; 2. One pair of indexes are the same: $i = j$ and $i' \neq j'$, or symmetrically $i \neq j$ and $i' = j'$; 3. No pair of indexes are the same: $i \neq j$ and $i' \neq j'$.

\noindent {\bf Part 1. Two pairs of indexes are the same.} We consider the case where $i = j$ and $i' = j'$.
We have
\begin{align}\label{eq:two_sketch_whp_part_1}
    \sum_{i=j,i'=j'} g_{i} h_{j'} \sigma_{i} \sigma_{j} \sigma'_{i'} \sigma'_{j'} \langle R_i, R_j \rangle A_{j,i'} \langle S_{i'}, S_{j'} \rangle = \sum_{i, i'} g_{i} h_{i'} A_{i,i'} = g^{\top} A h,
\end{align}
where the first step follows from $\langle R_i, R_i \rangle = \langle S_{i'}, S_{i'} \rangle = 1, \forall i, i' \in [n]$, see Eq.~\eqref{eq:two_sketch_same_Ri}.

\noindent {\bf Part 2. One pair of indexes are the same.} We consider the case where $i = j$ and $i' \neq j'$, or the symmetric case where $i \neq j$ and $i' = j'$.

W.l.o.g. we consider the case that $i = j$ and $i' \neq j'$. We have
\begin{align*}
    \sum_{i=j, i'\neq j'} g_{i} h_{j'} \sigma_{i} \sigma_{j} \sigma'_{i'} \sigma'_{j'} \langle R_i, R_j \rangle A_{j,i'} \langle S_{i'}, S_{j'} \rangle 
    = &~ \sum_{i, i'\neq j'} g_{i} h_{j'} \sigma'_{i'} \sigma'_{j'} A_{i,i'} \langle S_{i'}, S_{j'} \rangle  \\
    = &~ \sum_{j'} \sigma'_{j'} h_{j'} \sum_{i'\neq j'} \sigma'_{i'} (A^{\top} g)_{i'} \langle S_{i'}, S_{j'}\rangle,
\end{align*}
where the first step follows from $\langle R_i, R_i \rangle = 1, \forall i \in [n]$ (Eq.~\eqref{eq:two_sketch_same_Ri}), the second step follows from $\sum_i g_i A_{i,i'} = (A^{\top} g)_{i'}$.

Using Khintchine's inequality (Fact~\ref{fac:khintchine}) and Union bound, we have that with probability at least $1 - \poly(1/n)$,
\begin{align*}
    \Big(\sum_{j'} \sigma'_{j'} h_{j'} \sum_{i'\neq j'} \sigma'_{i'} (A^{\top} g)_{i'} \langle S_{i'}, S_{j'}\rangle \Big)^2
    \leq &~ O(\log n) \sum_{j'} h_{j'}^2 \Big(\sum_{i'\neq j'} \sigma'_{i'} (A^{\top} g)_{i'} \langle S_{i'}, S_{j'} \rangle \Big)^2 \\
    \leq &~ O(\log^2 n) \sum_{j'} h_{j'}^2 \sum_{i' \neq j'} (A^{\top} g)_{i'}^2 \langle S_{i'}, S_{j'}\rangle^2 \\
    \leq &~ O((\log^3 n)/b_2) \sum_{j'} h_{j'}^2 \sum_{i' \neq j'} (A^{\top} g)_{i'}^2 \\
    \leq &~ O((\log^3 n)/b_2) \|h\|_2^2 \|A^{\top} g\|_2^2,
\end{align*}
where the first step follows from applying Khintchine's inequality with $t=O(\sqrt{\log n})$, the second step again follows from applying Khintchine's inequality with $t=O(\sqrt{\log n})$, the third step follows from that with probability at least $1-\poly(1/n)$, $\langle S_i, S_j\rangle \leq O(\sqrt{(\log n) / b_2})$ for all $i\neq j \in [n]$, see Eq.~\eqref{eq:two_sketch_diff_Ri}. 

Plugging this equation into the previous equation, and note that the case that $i'=j', i\neq j$ is symmetric, we have that with probability at least $1 - \poly(1/n)$,
\begin{align}\label{eq:two_sketch_whp_part_2}
    &~ \sum_{\substack{i=j, i'\neq j' \\\mathrm{or~} i'=j', i\neq j}} g_{i} h_{j'} \sigma_{i} \sigma_{j} \sigma'_{i'} \sigma'_{j'} \langle R_i, R_j \rangle A_{j,i'} \langle S_{i'}, S_{j'} \rangle \\
    \leq &~ O(\log^{1.5} n / \sqrt{b_1}) \|g\|_2 \|A h\|_2 + O(\log^{1.5} n / \sqrt{b_2}) \|g^{\top} A\|_2 \|h\|_2.
\end{align}

\noindent {\bf Part 3. No pair of indexes are the same.} We consider the case where $i \neq j$ and $i' \neq j'$. We prove it by using Khintchine's inequality (Fact~\ref{fac:khintchine}) four times. We have that with probability $1 - \poly(1/n)$,
\begin{align*}
    &~ \Big( \sum_{i \neq j, i' \neq j'} g_{i} h_{j'} \sigma_{i} \sigma_{j} \sigma'_{i'} \sigma'_{j'} \langle R_i, R_j \rangle A_{j,i'} \langle S_{i'}, S_{j'} \rangle \Big)^2 \\
    = &~ \Big( \sum_{i} \sigma_i g_i \sum_{j'} \sigma'_{j'} h_{j'} \sum_{i' \neq j'} \sigma'_{i'} \langle S_{i'}, S_{j'} \rangle \sum_{j \neq i} \sigma_i \langle R_i, R_j \rangle A_{j,i'} \Big)^2 \\
    \leq &~ O(\log n) \sum_{i} g_i^2 \Big( \sum_{j'} \sigma'_{j'} h_{j'} \sum_{i' \neq j'} \sigma'_{i'} \langle S_{i'}, S_{j'} \rangle \sum_{j \neq i} \sigma_i \langle R_i, R_j \rangle A_{j,i'} \Big)^2 \\
    \leq &~ O(\log^2 n) \sum_{i} g_i^2 \sum_{j'} h_{j'}^2 \Big( \sum_{i' \neq j'} \sigma'_{i'} \langle S_{i'}, S_{j'} \rangle \sum_{j \neq i} \sigma_i \langle R_i, R_j \rangle A_{j,i'} \Big)^2 \\
    \leq &~ O(\log^3 n) \sum_{i} g_i^2 \sum_{j'} h_{j'}^2 \sum_{i' \neq j'} \langle S_{i'}, S_{j'} \rangle^2 \Big( \sum_{j \neq i} \sigma_i \langle R_i, R_j \rangle A_{j,i'} \Big)^2 \\
    \leq &~ O(\log^4 n) \sum_{i} g_i^2 \sum_{j'} h_{j'}^2 \sum_{i' \neq j'} \langle S_{i'}, S_{j'} \rangle^2 \sum_{j \neq i} \langle R_i, R_j \rangle^2 A_{j,i'}^2 \\
    \leq &~ O((\log^6 n) / (b_1 b_2)) \|g\|_2^2 \|h\|_2^2 \|A\|_F^2,
\end{align*}
where the second step follows from Khintchine's inequality with $t = O(\sqrt{n})$, the third step follows from Khintchine's inequality with $t = O(\sqrt{n})$ for each $i \in [n]$, and combining the $n$ inequalities using Union bound, the fourth step and the fifth step follows from same reason as the third step, the sixth step follows from that with probability at least $1-\poly(1/n)$, $\langle S_{i'}, S_{j'} \rangle \leq O(\sqrt{(\log n) / b_2})$ for all $i' \neq j' \in [n]$, and similarly with probability at least $1-\poly(1/n)$, $\langle R_i, R_j\rangle \leq O(\sqrt{(\log n) / b_1})$ for all $i\neq j \in [n]$, we combine the $2 n^2$ such bounds all $i,j,i',j' \in [n]$ using Union bound.

Thus we have that with probability at least $1 - \poly(1/n)$,
\begin{align}\label{eq:two_sketch_whp_part_3}
    \sum_{i \neq j, i' \neq j'} g_{i} h_{j'} \sigma_{i} \sigma_{j} \sigma'_{i'} \sigma'_{j'} \langle R_i, R_j \rangle A_{j,i'} \langle S_{i'}, S_{j'} \rangle
    \leq &~ O((\log^3 n) / \sqrt{b_1 b_2}) \cdot \|g\|_2 \|h\|_2 \|A\|_F.
\end{align}

\noindent {\bf Combining all parts together.} Adding Eq.~\eqref{eq:two_sketch_whp_part_1}, \eqref{eq:two_sketch_whp_part_2}, \eqref{eq:two_sketch_whp_part_3} together and plugging into Eq.~\eqref{eq:two_sketch_close_form_whp}, using Union bound, with probability at least $1 - \poly(1/n)$ we have,
\begin{align*}
    &~ g^{\top} (R^{\top} R) A (S^{\top} S) h - g^{\top} A h \\
    \leq &~ O(\frac{\log^{1.5} n}{\sqrt{b_1}}) \|g\|_2 \|A h\|_2 + O(\frac{\log^{1.5} n}{\sqrt{b_2}}) \|g^{\top} A\|_2 \|h\|_2 + O(\frac{\log^{3} n}{\sqrt{b_1 b_2}}) \cdot \|g\|_2 \|h\|_2 \|A\|_F.
\end{align*}
\end{proof}

%% file: exp.tex
\section{Experiments}\label{sec:exp}

\begin{table*}[!t]
\centering
\resizebox{\textwidth}{!}{
\begin{tabular}{|l|c|c|c|c|c|c|c|} \toprule 
\textbf{Datasets} & COLLAB & IMDB-B & IMDB-M & PTC & NCI1 & MUTAG & PROTEINS \\\midrule
\# of graphs & $5000$ & $1000$ & $1500$ & $344$ & $4110$ & $188$ & $1113$ \\[0.7ex]
\# of classes & $3$ & $2$ & $3$ & $2$ & $2$ & $2$ & $2$ \\[0.7ex]
Avg \# of nodes & $74.5$ & $19.8$ & $13.0$ & $25.5$ & $29.8$ & $17.9$ & $39.1$ \\[0.7ex] \midrule \\[-1.8ex]
GNTK & $>24$ hrs & $546.4$ & $686.0$ & $46.5$ & $10,084.7$ & $8.0$ & $1,392.0$ \\[0.7ex] \midrule \\[-1.8ex]
\textbf{Ours} & $4,523.0 \ (\boldsymbol{>19 \times})$ & $90.7 \ (\boldsymbol{6 \times})$ & $112.5 \ (\boldsymbol{6.1 \times})$ & $13.5 \ (\boldsymbol{3.4 \times})$ & $7,446.8 \ (\boldsymbol{1.4 \times})$ & $3.0 \ (\boldsymbol{2.7 \times})$ & $782.7 \ (\boldsymbol{1.8 \times})$ \\[0.7ex]
\bottomrule
\end{tabular}} 
\caption{ Running time comparison between our method and GNTK. All the numbers are in seconds.
}
\label{tab:acc}
\end{table*}

\paragraph{Datasets.}
We test our method on 7 benchmark graph classification datasets, including 3 social networking dataset (COLLAB, IMDBBINARY, IMDBMULTI) and 4 bioinformatics datasets (PTC, NCL1, MUTAG and PROTEINS) \cite{ypv15}. For bioinformatics dataset, each node has its categorical features as input feature. For each social network dataset where nodes have no input feature, we use the degree of each node as its feature to represent its structural information. The dataset statistics are shown in Table \ref{tab:acc}.

\paragraph{Results.}
We compare our performance with original GNTK \cite{dhs+19}. The results are shown in Table \ref{tab:acc}. Our matrix decoupling method (MD) doesn't change the result of GNTK while significantly accelerates the learning time of neural tangent kernel. Our proposed method achieves multiple times of improvements for all the datasets. In particular, on COLLAB, our method achieves more than $19$ times of learning time acceleration. We observe that the improvement of our method depends on the sizes of the graphs. For large-scale dataset like COLLAB, we achieve highest acceleration because matrix multiplication dominates the overall calculation time. For bioinformatics datasets where number of nodes is relatively small, the improvement is not as prominent.